\pdfoutput=1

\documentclass[11pt]{article}

\usepackage[final]{acl}

\usepackage{times}
\usepackage{latexsym}

\usepackage[T1]{fontenc}

\usepackage[utf8]{inputenc}

\usepackage{microtype}

\usepackage{inconsolata}

\usepackage{amsmath}
\usepackage{amsthm}
\usepackage{amsfonts}
\usepackage{multirow}
\usepackage{booktabs}
\usepackage{multirow}
\usepackage{algorithm}
\usepackage{algorithmic}
\usepackage{graphicx}
\usepackage[table]{xcolor}
\title{1+1\(>\)2: A Synergistic Sparse and Low-Rank Compression Method for Large Language Models}

\author{Zeliang Zong$^{1}\thanks{{ }{ }Equal contribution.}$, 
\textbf{Kai Zhang}$^{1 *}$, 
\textbf{Zheyang Li}$^{1}$, 
\textbf{Wenming Tan}$^{1}$, 
\textbf{Ye Ren}$^{1}$\thanks{{ }{ }Corresponding author.}, 
\\
\textbf{Yiyan Zhai}$^{1}$\textbf{,}
\textbf{Jilin Hu}$^{2}$
\\
$^{1}$ Hikvision Research Institute \\
$^{2}$ School of Data Science and Engineering, East China Normal University \\
\texttt{\{zongzeliang, zhangkai, lizheyang, tanwenming, renye\}@hikvision.com} \\
\texttt{zhaiyiyan@163.com, jlhu@dase.ecnu.edu.cn}
}

\begin{document}
\maketitle
\begin{abstract}
Large Language Models (LLMs) have demonstrated remarkable proficiency in language comprehension and generation; however, their widespread adoption is constrained by substantial bandwidth and computational demands. While pruning and low-rank approximation have each demonstrated promising performance individually, their synergy for LLMs remains underexplored. We introduce \underline{S}ynergistic \underline{S}parse and \underline{L}ow-Rank \underline{C}ompression (SSLC) methods for LLMs, which leverages the strengths of both techniques: low-rank approximation compresses the model by retaining its essential structure with minimal information loss, whereas sparse optimization eliminates non-essential weights, preserving those crucial for generalization. Based on theoretical analysis, we first formulate the low-rank approximation and sparse optimization as a unified problem and solve it by iterative optimization algorithm. Experiments on LLaMA and Qwen2.5 models (7B-70B) show that SSLC, without any additional training steps, consistently surpasses standalone methods, achieving state-of-the-arts results. Notably, SSLC compresses Qwen2.5 by 50\% with no performance drop and achieves at least 1.63$\times$ speedup, offering a practical solution for efficient LLM deployment.
\end{abstract}

\section{Introduction}

\label{sec:intro}

\begin{figure}[t]
    \centering
    \includegraphics[width=0.95\linewidth]{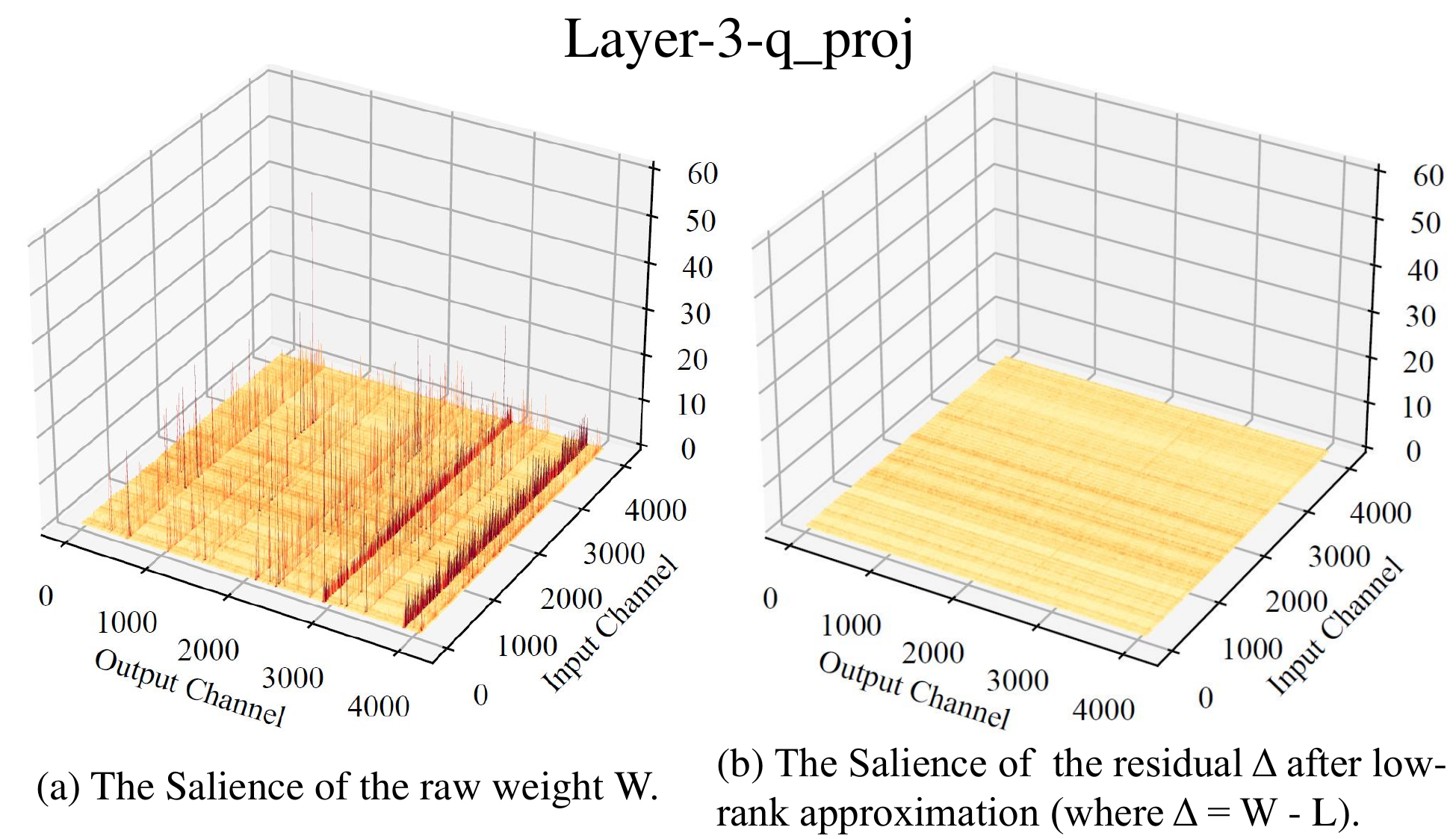}
    \caption{Weight salience~\cite{huang2024slim} in LLaMA2-7B before and after synergistic low-rank approximation. Compared to Figure (a), Figure (b) not only shows a substantial reduction in extreme high values, but also reveals a decrease in prunable low values, thus mitigating the performance degradation caused by pruning.}
    \label{fig:salience_distribution_3D}
\end{figure}

 In the research field of natural language processing (NLP), large language models (LLMs)~\cite{zhang2022opt,scao2022bloom,touvron2023llama}, as an emerging technology, have achieved remarkable success in handling complex linguistic tasks and have significantly influenced the evolutionary direction of NLP~\cite{bubeck2023sparks,wei2022emergent,achiam2023gpt}. However, their vast parameters require extensive computational resources and substantial memory bandwidth, thereby constraining their deployment in practical applications.
 
To address the memory consumption issues of LLMs, various post-training compression (PTC) techniques that do not require retraining have been explored. These include model quantization~\cite{dettmers2022llmint8,xiao2022smoothquant,frantar2023gptq,liu2025spinquantllmquantizationlearned}, pruning~\cite{frantar2023sparsegpt,sun2023simple,ma2023llmv3} and low-rank approximation~\cite{hsu2022language,yuan2023asvd,wang2024svd}. Pruning simplifies the network by removing non-critical weights or structures, while low-rank approximation methods reduces the model's complexity by decomposing the weight matrix into two orthogonal low-dimensional matrices.

Recent studies~\cite{frantar2023sparsegpt,sun2023simple,zhang2024plug,dong2024pruner,meng2024alpsimprovedoptimizationhighly} have formulated LLM pruning as a layer-wise reconstruction problem and pruned redundant neurons using a metric derived from the second Taylor approximation of reconstruction error~\cite{hassibi1993obs}. This metric, referred to as weight salience~\cite{huang2024slim} and detailed in the preliminaries section, evaluates the quadratic error associated with changes in matrix elements, which directly correlates with model performance: higher salience indicate a greater impact on performance. As illustrated in Figure~\ref{fig:salience_distribution_3D}(a), the original weight salience, approximated from the calibration dataset that is conventionally employed by prevailing methodologies~\cite{frantar2023sparsegpt,sun2023simple}, exhibits a discrete distribution of outliers against a consistent pattern of moderate values. Unfortunately, existing pruning approaches retain neurons with high salience from a discrete perspective, failing to maximize the extraction of the coherent part in salience space. In contrast, low-rank approximation (LRA) methods, such as Singular Value Decomposition (SVD)~\cite{hsu2022language, yuan2023asvd,wang2024svd}, are particularly suitable for compressing the coherent components within the salience and extracting a set of orthogonal bases that form a subspace, maximizing the preservation of the energy of the original space. However, these methods for LLMs still lead to severe performance degradation at a high compression ratio~\cite{yuan2023asvd,wang2024svd}. This degradation arises because low-rank approximation effectively preserves the weight-sharing common basis, but fails to retain the full-rank, non-coherent parts that are crucial for maintaining the model’s knowledge and performance.

Given these insights, there is an urgent need to combine sparsification and low-rank approximation techniques. This integration can enhance compression efficiency while ensuring that critical information is preserved. Figure~\ref{fig:salience_distribution_3D} demonstrates that the outliers in salience space are effectively extracted after low-rank approximation, and this phenomenon is quantitatively analyzed in Section~\ref{sec:compresion_rate_spread}. Consequently, with the same compression rate, the synergistic method, by truncating at a smaller salience threshold and increasing the proportion of neurons with less salience, leads to fewer reconstruction errors and thus less performance degradation. 

Inspired by these experimental observations, we propose the Synergistic Sparse and Low-Rank Compression (SSLC) method. SSLC decouples the coherent and non-coherent parts of the neuron, allowing the model to benefit from both sparse and low-rank approximation. The low-rank approximation uses orthogonal bases to maximize the extraction of energy from the salience space, while the sparse part preserves key incoherent neurons to maintain the network's essential expressive power. By synergizing these two techniques, SSLC ensures a dense, expressive layer with the low-rank part, mitigating the loss of expressive capacity caused by pure pruning/sparsification. Furthermore, we model the joint compression problem as a unified data-aware mathematical optimization objective, considering the effect of low-rank and sparse components on reconstruction loss. Then, a synergistic optimization algorithm has been proposed to solve the problem. Consequently, our method possesses the orthogonality property of low-rank approximation and the full-rank property of sparsification mathematically, ensuring effective preservation of the model's expressive capacity while reducing redundant information. Another advantage, based on the assumption that weight changes during model adaptation exhibit a low ``intrinsic rank''~\cite{aghajanyan2020intrinsic,hu2021lora}, the low-rank component can effectively adapts to downstream tasks. Through comprehensive experiments on the LLaMA~\cite{touvron2023llama,touvron2023llama2,grattafiori2024llama3herdmodels} and Qwen2.5~\cite{qwen2025qwen25technicalreport} models with 7B to 70B parameters, the results demonstrate that SSLC achieves state-of-the-art performance.

The main contributions are summarized as follows: 
\begin{itemize}
    \item We propose SSLC, a novel joint compression algorithm that integrates low-rank approximation with pruning techniques. Mathematically, our method demonstrates the benefits of both orthogonality from low-rank approximation and full-rank preservation via sparse reconstruction.  
    \item Extensive experiments have shown that SSLC without fine-tuning achieves state-of-the-art performance on various models and datasets. In addition, SSLC provides an optimized initialization for subsequent low-rank part fine-tuning. Specifically, SSLC yields a 1.63× speedup on Qwen2.5-7B (within about 3 GPU hours of pruning and fine-tuning) without performance drop across various zero-shot tasks. 
\end{itemize}

\section{Related Works}\label{sec:related}

\subsection{Large Language Models Pruning}\label{sec:related:LLMpruning}
SparseGPT~\cite{frantar2023sparsegpt} pioneers LLM pruning using a metric derived from the second-order term in the Taylor expansion of the reconstruction error, employing classical Optimal Brain Surgeon (OBS) techniques~\cite{hassibi1992second} to iteratively prune the network and update residual weights. Wanda~\cite{sun2023simple} simplifies the Hessian matrix inversion process, focusing on pruning the smallest magnitudes multiplied by the corresponding input activation. RIA~\cite{zhang2024plug} introduces the Relative Importance and Activation metric and channel swapping to maximize the retention of salience under N:M sparsity constraints. DSNoT~\cite{zhang2024dynamic} iteratively prunes and grows weights to minimize reconstruction loss without the computational expense of back-propagation or weight updates. ALPS~\cite{meng2024alpsimprovedoptimizationhighly} utilizes an ADMM-based optimization framework to alternately optimize remaining weights through iterative closed-form updates, minimizing layer-wise reconstruction error while satisfying sparsity constraints. Pruner-Zero~\cite{dong2024pruner}, automatically generate symbolic pruning metrics, exploring correlations with post-pruning performance. These methods focus on model compression  purely from a pruning perspective. In contrast, our approach emphasizes the synergy between pruning and low-rank approximation, effectively minimizing the impact of pruning on reconstruction loss.

\subsection{Sparse and Low-Rank Integration}
Early joint decomposition research, including Robust Principal Component Analysis (RPCA)~\cite{wright2009robust} and GoDec~\cite{zhou2011godec}, effectively decoupled low-rank structures and sparse noise from data matrices.  LoSparse~\cite{li2023losparse} decomposes model weights into low-rank and sparse components via iterative pruning, yet remains impractical for LLMs due to full-network training demands. Techniques like LoRAshear~\cite{chen2023lorashear} and LoRAPrune~\cite{zhang2023loraprune} integrate pruning with LoRA, performing parameter pruning based on gradient information from LoRA, primarily designed for structured pruning, but still face challenges for severe performance degradation at a high compression ratio. Meanwhile, LoSA~\cite{huang2025dynamiclowranksparseadaptation} further enhances compressed LLM performance by unifying LoRA with sparsity optimization. Additionally, LoRaP~\cite{li2024lorap} applies separate low-rank estimation and pruning to MHA and MLP layers independently; however, it lacks joint optimization and requires additional LoRA branch fine-tuning during knowledge recovery, limiting its efficiency. In contrast to these paradigms that conditionally adapt Low-rank either for gradient approximation or fine-tuning, our SSLC framework pioneers a unified matrix-level decomposition where both low-rank and sparse components are jointly optimized via second-order reconstruction loss, enabling data-aware compression and direct mining of latent low-rank representations to drive efficient compression.

\begin{figure*}[t]
    \centering
    \includegraphics[width=0.86\linewidth]{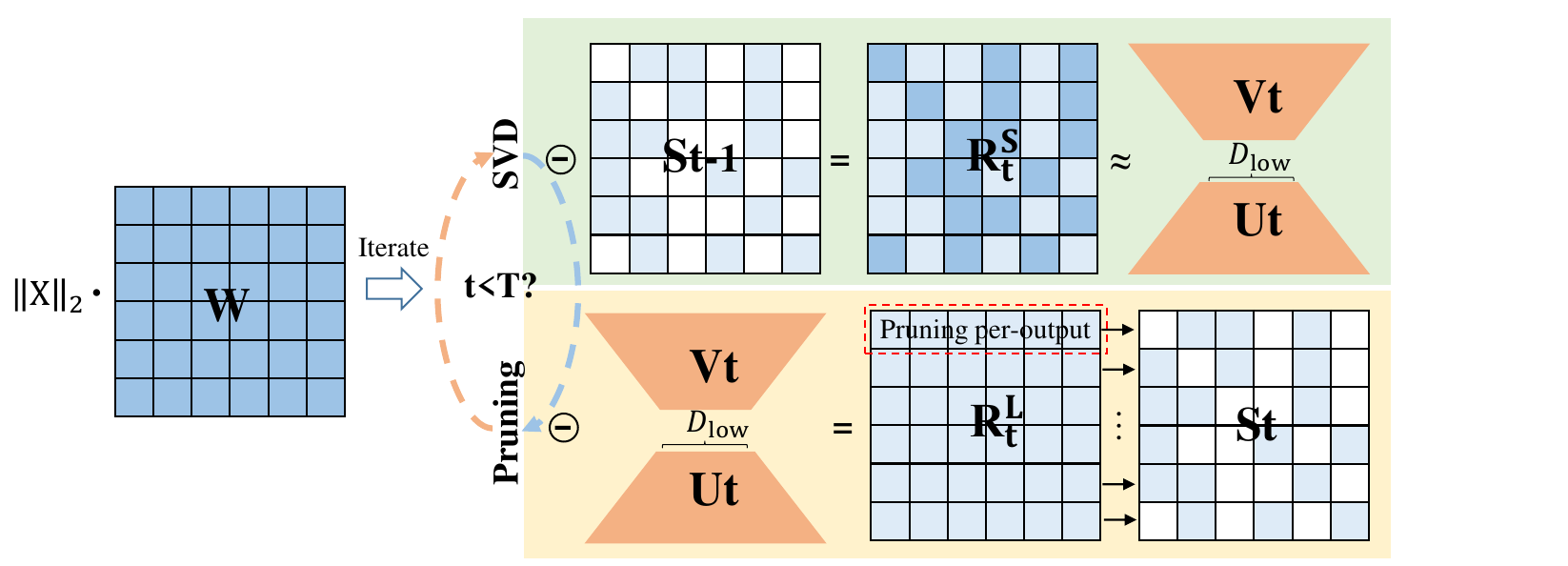}
    \caption{The pipeline of our proposed SSLC method involves the following steps: Initially, the SVD step performs a low-rank approximation on the scaled matrix. Subsequently, the pruning step converts the dense matrix into a sparse one. In essence, SSLC executes \(T\)-step SVD and pruning iterations on the scaled matrix, decomposing the original weight matrix W into a sparse matrix \(S_t\) and low-dimensional matrices \(V_t\) and \(U_t\). After the final iteration, the method multiplies \(V_t\) and \(S_t\) by the scaling matrix \( \left \|X \right \| _2^{-1}\), to revert to the original matrix state before scaling.}
    \label{fig:overview}
     \vspace{-0.5cm}
\end{figure*}

\section{Preliminaries}\label{sec:pre}
Current post-training compression methods focus on compressing pre-trained weights without retraining, ensuring model performance by minimizing the output discrepancy between the compressed and original models. Due to the computational infeasibility of global minimization, this task is typically framed as a layer-wise reconstruction problem for LLMs. Let \(W\in \mathbb{R}^{(m,n)}\) and \(W^{'}\in\mathbb{R}^{(m,n)}\) denote the original and compressed weights of a given layer, where \(m\) and \(n\) represent the number of output and input channels, respectively. The input activation is represented as \(X\in\mathbb{R}^{(n,N\times L)}\), where \(N\) is the number of calibration samples and \(L\) is the sequence length respectively. This problem can be expressed as follows:
\begin{equation}
    \arg\min_{W^{'}}\left \| (W- W^{'})X\right \| _F
    \label{eq:costfun}
\end{equation}
where \(\left \|\cdot \right \| _F\) is the Frobenius norm. To prune or quantize weights with minimal impact on the optimization objective, rigorous mathematical derivations from works such as Optimal Brain Surgeon (OBS)~\cite{hassibi1992second} and Optimal Brain Quantization (OBQ)~\cite{frantar2022optimal}, as well as applications like SparseGPT~\cite{frantar2023sparsegpt} and GPTQ~\cite{frantar2023gptq} on LLMs, suggest that the change of the element at \((i, j)\) induces a quadratic error to the cost function Eq.~\ref{eq:costfun}. Specifically, the error \(\delta_{i,j}\) is approximated by: \(\frac{\Delta W_{ij}^{2} }{[H^{-1}]_{j,j}^2}  \). The Hessian matrix is approximated as \(H\approx X^{T}X\) for a weight matrix. For instance, in quantization, \(\Delta w_{ij}=w_{ij}-quant(w_{ij})\); in pruning, \(\Delta w_{ij}=w_{ij}-0\). Here, \([H^{-1}]_{j,j}^2\) denotes the \(j\)-th diagonal entry of the inverse Hessian matrix.

\section{Method}\label{sec:method}

The section presents our proposed method, Synergistic Sparse and Low-Rank Compression (SSLC) for LLMs, as illustrated in Figure~\ref{fig:overview}. The method comprises three principal sections: the proposed low-rank aware optimization objective, the synergistic optimization algorithm, and the process of low-rank fine-tuning recovery.

\subsection{Joint Low-rank and Sparse Compression}\label{sec:method:optimization_objective}

Low-rank decomposition and pruning methods based solely on weight magnitudes have been shown empirically ineffective~\cite{frantar2023sparsegpt,yuan2023asvd}. Unlike existing methods~\cite{li2023loftq} that directly decompose a matrix \(W\), our method employs a data-aware synergistic optimization strategy. We decompose the original outputs into a low-rank part \(L\in \mathbb{R}^{(m,n)}\) with rank \(r\) and a sparse part \(S\in \mathbb{R}^{(m,n)}\) with sparsity \(k\%\), minimizing the following objective:
\begin{equation}
\begin{array}{c}
\min_{L, S} \lVert (W - L - S)X \rVert_F \\ 
\text{s.t. } \operatorname{rank}(L) = r,\ \operatorname{sparsity}(S) = k\%
\end{array}
\label{eq:ob}
\end{equation}
The functions \(\text{rank}(\cdot)\) and \(\text{sparsity}(\cdot)\) are used to obtain the rank and sparsity of a matrix, respectively. This optimization objective jointly accounts for the contributions of both low-rank and sparse components to output reconstruction loss. In contrast, prior approaches optimize only one aspect—either designing better pruning metrics or singular values mapped to the objective—while ignoring the synergistic benefits of combining both.

\subsection{Synergistic Optimization Algorithm}\label{sec:method:synergistic_optimization}

Unlike RPCA~\cite{wright2009robust} which decomposes data matrices into low-rank and sparse components based on pure mathematical objectives, SSLC introduces data-awareness through layer-wise reconstruction error minimization, explicitly aligning decomposition with LLM performance preservation. Decomposing a low-rank matrix and a sparse matrix simultaneously from Eq.~\ref{eq:ob} is a NP-
hard problem. To facilitate the synergistic optimization, we break down the optimization problem into two manageable sub-problems, enabling efficient alternation between sparsification and singular value decomposition (SVD): 
\begin{equation}
\left\{
\begin{array}{c}
   
S_t =\mathop{\arg\min}\limits_{\operatorname{sparsity}(S) = k\%} \left\| (W - L_t - S)X \right\|_F \\
L_t= \mathop{\arg\min}\limits_{\operatorname{rank}(L) = r}\left\| (W - L - S_{t-1})X \right\|_F

\end{array}
\right.
\label{eq:osp}
\end{equation}
Here, \(L_t\) and \(S_t\) denote the low-rank and sparse matrices at the \(t\)-th iteration step, respectively. 

\subsubsection{Sparsification} When solving for the sparse matrix in Eq.~\ref{eq:osp} at the \(t\)-th iteration, the low-rank matrix \(L_t\) is computed in advance, allowing us to sparsify the residual of the low-rank approximation (\(R_t^{L}=W-L_t\)). Nevertheless, directly solving for the binary mask corresponding to the weight matrix of LLM using a differentiable approach is impractical due to the immense size of the solution space. Recently,  Methods~\cite{frantar2023sparsegpt,sun2023simple,zhang2024dynamic} following OBD~\cite{lecun1990obd} and OBS~\cite{hassibi1993obs} has gained traction in the field of LLM pruning, which use calibration data to select the most salient weights and to minimize block reconstruction errors effectively. The salience ($\delta$) of residual weights for pruning is approximated as follows:

\begin{equation}
\begin{aligned}
\delta_{ij}&=\left [  \left | R_t^{L} \right |^{2}/diag\left (  \left (  X^{T}X\right )^{-1} \right )  \right ]_{ij} \\ &\underset{approx.}{\overset{diagonal}{=}}  \left ( \left | R_t^{L} \right |\cdot \left \| X_j \right \|_2   \right )_{ij}^2 
\end{aligned}
\label{eq:prune}
\end{equation}
Then, the residual matrix are pruning according to \(\theta\), which is the \textit{k}-th percentile of the sorted salience in descending order.

\begin{equation}
[S_t]_{ij}=\left\{\begin{matrix}[R^S_t]_{ij} & \text{if}~\delta_{ij} \geq \theta
 \\
0 & \text{otherwise} 
\end{matrix}\right.
\label{eq:prune1}
\end{equation}


\begin{table*}[ht!]
    \centering
    \renewcommand{\arraystretch}{1}
    \resizebox{\linewidth}{!}{
    \begin{tabular}{ccccccccccc}
\toprule
\multirow{2}{*}{Task} & \multirow{2}{*}{Methods} & \multirow{2}{*}{Type} & \multicolumn{6}{c}{LLaMA} & \multicolumn{2}{c}{Qwen2.5} \\ 
                                \cmidrule(lr){4-11}
 & & & 1-7B & 2-7B & 3-8B & 1-13B & 2-13B & 3-70B & 7B & 14B 
 \\ \hline
                            & Dense                                 & -                                      & 7.34                                  & 7.26                                  & 9.54                                   & 6.70                                  & 6.73                                  & 7.17                                  & 11.86                                  & 10.35                                  \\ \cline{2-11} 
                            & SparseGPT                             & S                                      & 9.31                                  & 9.23                                  & 14.25                                  & 8.12                                  & 8.22                                  & 9.66                                  & 13.89                                  & 12.41                                  \\
                            & Wanda                                 & S                                      & 9.30                                  & 9.24                                  & 14.87                                  & 8.13                                  & 8.30                                  & 9.96                                  & 14.24                                  & 12.40                                  \\
                            & DSnoT                                 & S                                      & 9.13                                  & 9.11                                  & 14.58                                  & 8.06                                  & 8.13                                  & 9.92                                  & 14.19                                  & 12.23                                  \\
                            & SVD-LLM                               & LRA                                    & 127.25                                & 161.27                                & 413.74                                 & 53.41                                 & 87.20                                 & \multicolumn{1}{c}{154.19}            & 379.64                                 & 307.18                                 \\
\multirow{-6}{*}{C4}        & \cellcolor[HTML]{EFEFEF}\textbf{Ours} & \cellcolor[HTML]{EFEFEF}\textbf{S+LRA} & \cellcolor[HTML]{EFEFEF}\textbf{8.91} & \cellcolor[HTML]{EFEFEF}\textbf{8.87} & \cellcolor[HTML]{EFEFEF}\textbf{13.90} & \cellcolor[HTML]{EFEFEF}\textbf{7.91} & \cellcolor[HTML]{EFEFEF}\textbf{8.02} & \cellcolor[HTML]{EFEFEF}\textbf{9.39} & \cellcolor[HTML]{EFEFEF}\textbf{13.59} & \cellcolor[HTML]{EFEFEF}\textbf{12.02} 
 \\ \hline
                            & Dense                                 & -                                      & 5.68                                  & 5.47                                  & 6.24                                   & 5.09                                  & 4.88                                  & 2.86                                  & 6.85                                   & 5.29                                   \\ \cline{2-11} 
                            & SparseGPT                             & S                                      & 7.22                                  & 6.99                                  & 9.29                                   & 6.21                                  & 6.02                                  & 5.77                                  & 8.43                                   & 7.28                                   \\
                            & Wanda                                 & S                                      & 7.24                                  & 6.92                                  & 9.65                                   & 6.15                                  & 5.97                                  & 5.82                                  & 8.62                                   & 7.32                                   \\
                            & DSnoT                                 & S                                      & 7.15                                  & 6.84                                  & 9.52                                   & 6.09                                  & 5.87                                  & 5.79                                  & 8.58                                   & 7.23                                   \\
                            & SVD-LLM                               & LRA                                    & 24.52                                 & 27.82                                 & 42.63                                  & 13.71                                 & 15.76                                 & \multicolumn{1}{c}{12.65}             & 38.64                                  & 26.13                                  \\
\multirow{-6}{*}{Wiki2} & \cellcolor[HTML]{EFEFEF}\textbf{Ours} & \cellcolor[HTML]{EFEFEF}\textbf{S+LRA} & \cellcolor[HTML]{EFEFEF}\textbf{6.92} & \cellcolor[HTML]{EFEFEF}\textbf{6.61} & \cellcolor[HTML]{EFEFEF}\textbf{8.95}  & \cellcolor[HTML]{EFEFEF}\textbf{5.96} & \cellcolor[HTML]{EFEFEF}\textbf{5.79} & \cellcolor[HTML]{EFEFEF}\textbf{5.36} & \cellcolor[HTML]{EFEFEF}\textbf{8.36}  & \cellcolor[HTML]{EFEFEF}\textbf{7.11}  \\ \hline
                            & Dense                                 & -                                      & 66.31                                  & 66.96                                  & 71.41                                  & 68.91                                  & 69.95                                  & 76.91                                  & 70.83                                  & 73.93                                  \\ \cline{2-11}
                            & SparseGPT                             & S                                      & 63.12                                  & 63.71                                  & 65.44                                  & 65.98                                  & 67.22                                  & 74.19                                  & 67.81                                  & 71.19                                  \\
                            & Wanda                                 & S                                      & 62.77                                  & 64.13                                  & 65.51                                  & 66.58                                  & 68.01                                  & 74.39                                  & 66.70                                  & 71.15                                  \\
                            & DSnoT                                 & S                                      & 62.91                                  & 63.22                                  & 64.91                                  & 66.41                                  & 67.78                                  & 74.27                                  & 66.89                                  & 71.23                                  \\
                            & SVD-LLM                                 & LRA                                      & 39.07                                  & 38.13                                  & 36.65                                  & 43.12                                  & 39.32                                 & 44.86                                  & 36.11                                  & 40.77                                  \\
\multirow{-5}{*}{\begin{tabular}[c]{@{}c@{}}Zero-\\shot  \end{tabular}} & \cellcolor[HTML]{EFEFEF}\textbf{Ours} & \cellcolor[HTML]{EFEFEF}\textbf{S+LRA} & \cellcolor[HTML]{EFEFEF}\textbf{63.59} & \cellcolor[HTML]{EFEFEF}\textbf{65.24} & \cellcolor[HTML]{EFEFEF}\textbf{65.97} & \cellcolor[HTML]{EFEFEF}\textbf{66.99} & \cellcolor[HTML]{EFEFEF}\textbf{68.55} & \cellcolor[HTML]{EFEFEF}\textbf{74.79} & \cellcolor[HTML]{EFEFEF}\textbf{68.68} & \cellcolor[HTML]{EFEFEF}\textbf{71.93} \\ \hline
\end{tabular}
    }
    \caption{Performance comparison of unstructured compression methods on LLaMA \& Qwen2.5 (50\% parameters remaining) without finetuning across three task categories: (S means Sparsification; C4 \& Wiki2 [WikiText-2] evaluated by perplexity [$PPL\downarrow$]; Zero-shot tasks reported as accuracy [\%] averaged over \{HellaSwag, Winogrande, BoolQ, PIQA, ARC-Easy, ARC-Challenge\}), with detailed per-dataset results in Appendix ~\ref{sec:detail_performance}.} 
    \label{tab:language_modelling}
    \vspace{-1mm}
\end{table*}

\subsubsection{SVD} After obtaining the sparse matrix, the sparse residual \(R_t^{S}=W-S_{t-1}\) can be calculated, the SVD sub-problem now be \(L_t= \mathop{\arg\min}\limits_{\text{rank}(L) = r}\left\| (R_t^S - L )X \right\|_F\). Although the SVD sub-problem can be directly solved by means of closed-form solutions as presented in~\cite{xiang2012optimal,saha2024compressinglargelanguagemodels}, the computational burden of performing two full SVD for large-scale matrices, such as those of dimensions \(4096\times4096\) and \(4096\times11008\), during the iterative process is prohibitively high. Accordingly, by referring to Section~\ref{sec:pre} and Eq.~\ref{eq:prune}, the impact of weight changes on the reconstruction loss following SVD compression can approximated efficiently. To minimize this impact, we construct a matrix that multiplies \(L_t'\) with rank \(r\) by the inverse of \(||X||^2\) as part of low-rank approximation. The optimization objective of this sub-problem can be approximated in the following form:
\begin{equation}
\begin{aligned}
       L_t'&=\arg\min_{L_t'}  \sum \left (  \left | R_t^{S}-L_t'\cdot||X||_2^{-1} \right |\cdot\left \|  X\right \|_2 \right )^{2} \\ 
          &=\arg\min_{L_t'}  \sum \left ( \left |R_t^{S}\cdot\left \|  X\right \|_2-L_t'  \right |  \right )^2  
\end{aligned}
\end{equation}

Hence, to improve efficiency while maintaining performance, a randomized SVD approach is adopted~\cite{zhou2011godec}. After applying randomized SVD for \(R_t^S\cdot\left \|  X\right \|_2\) , we obtain \(L_t '\). \(L_t '\) is represented as:
\begin{equation}
\begin{aligned}
 \tilde{L} &= R_t^S\cdot\left \|  X\right \|_2;\\
 Y_1&= \tilde{L}A_1, Y_2=\tilde{L}^TA_2;\\
 L_t'&=Y_1\left (A_2^TY_1 \right )^{-1} Y_2^T
\end{aligned}
\label{eq:svd}
\end{equation}
Obtaining \(Y_1\) and \(Y_2\) as the bilateral random projections (BRP) of matrix \(\tilde{L}\) through the application of random matrices \(A_1\) and \(A_2\), where \(A_1\in\mathbb{R}^{(n,r)}\) and  \(A_2\in\mathbb{R}^{(r,m)}\). Consequently, the two sub-problem within Eq.\ref{eq:osp} can be resolved efficiently as delineated below: 
\begin{equation}
\left\{
\begin{array}{c}
[S_t]_{ij}=\left\{\begin{matrix}[R^S_t]_{ij} & \text{if}~\delta_{ij} \geq \theta
 \\
0 & \text{otherwise} 
\end{matrix}\right. \\
L_t=L_t'\cdot\left \|  X\right \|_2^{-1}=Y_1\left (A_2^TY_1 \right )^{-1} Y_2^T\cdot\left \|  X\right \|_2^{-1}
\end{array}
\right.
\label{eq:solutions}
\end{equation}

\subsubsection{Preserving Most Important Weights} Recognizing the importance of the top significant weights ~\cite{dettmers2023spqr,yuanpb,huang2024slim}, we preserve the top 1\% of weights with highest salience (Eq.~\ref{eq:prune}) and exclude them from the synergistic decomposition process. To achieve an overall compression rate of \(p\%\), we allocate \((k-1)\%\) to the sparse part and \(r\times \frac{m+n}{m\times n} \) to the low-rank part, ensuring the sum of these proportions and the top 1\% preserved parameters equates \(p\%\). 

Optimizing each matrix independently allows for parallel execution, enhancing computational efficiency. Throughout the iteration process, we maintain the column norm \(||X||^2\) of the input vectors constant, while updating the residual matrices \(R_t^{S}\) and \(R_t^{L}\) dynamically. The overall algorithmic flow is depicted in Algorithm~\ref{alg:algorithm}. 

\begin{algorithm}[t]
\caption{SSLC Algorithm}
\label{alg:algorithm}
\textbf{Input}: Pre-trained weight matrix  \( W \) with the top 1\% significant values preserved\\
\textbf{Parameter}: Target rank \( r \), target sparsity \( (k-1)\% \), sparse algorithm \( \text{Sparse}(\cdot) \), alternating step \( T \) \\
\textbf{Output}: Sparse and low rank matrix \( S_t,L_t\)

\begin{algorithmic}[1] 
\STATE Let \(S_0=0\).
\FOR{\(t = 1\ to\ T\)}   
\STATE Obtain \(L_t \leftarrow \text{SVD}(W-S_{t-1}, r)\) by Eq.\ref{eq:svd}
\STATE Obtain \(S_t \leftarrow \text{Sparse}(W-L_t, (k-1)\%)\) by Eq.\ref{eq:prune}
\STATE \(t=t+1\)
\ENDFOR
\STATE \textbf{return} solution
\end{algorithmic}
\end{algorithm}

\subsection{Low-rank Fine-tuning Recovery}\label{sec:method:recovery}

\begin{figure}[ht]
    \centering
    \includegraphics[width=1.05\linewidth]{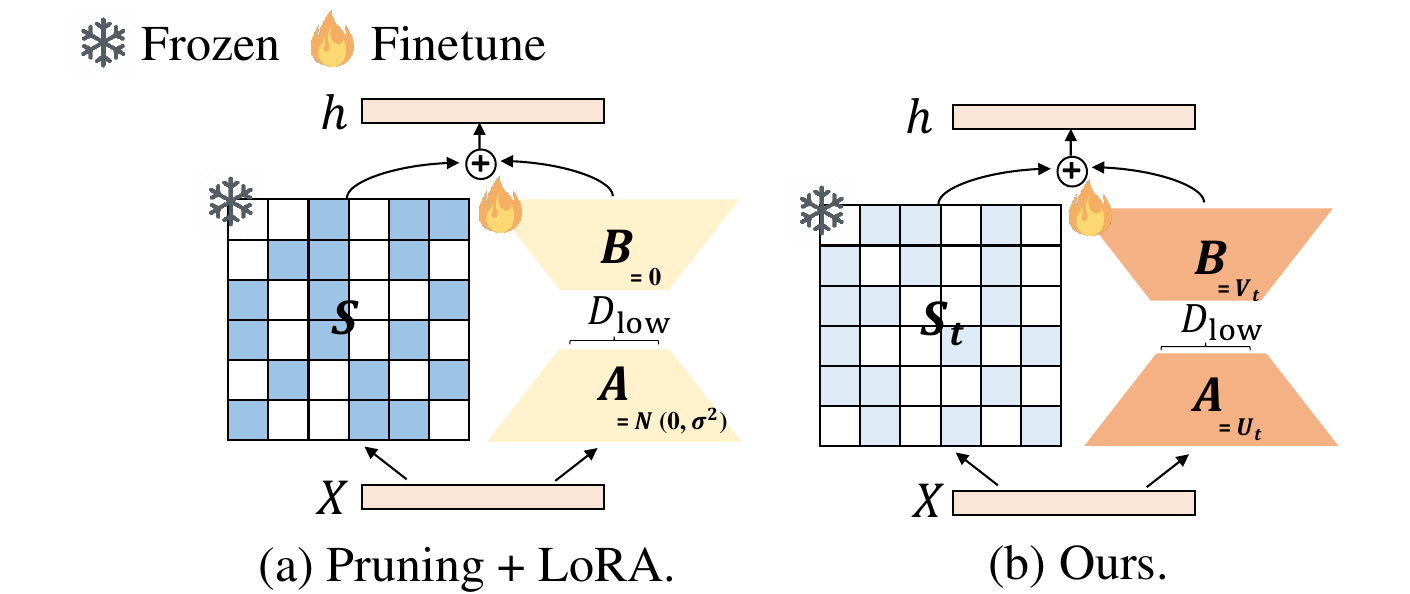}
    \caption{Fine-tuning under different types of pruning. (a) introduces an additional LoRA parameter. In contrast, the low-dimensional matrix ($D_{low} \leq 128 $) from SSLC framework can be directly used for fine-tuning.}
    \label{fig:fine-turning}
     \vspace{-0.5cm}
\end{figure}

Instead of directly inserting LoRA side, we use the \(U_t\) and \(V_t\) matrices decomposed from \(L_t\) for performance recovery. This approach maintains the sparse matrix \(S_t\) frozen and updates only the \(U_t\) and \(V_t\) matrices during fine-tuning, as shown in Figure~\ref{fig:fine-turning}, which can be expressed as:
\begin{equation}
\begin{aligned}
h &= (U_tV_t^{T}+S_t+\Delta W)X+b \\ &=({U_t}^{\prime}{V_t}^{T\prime}+S_t)X+b\label{eq5}
\end{aligned}
\end{equation}
where \(h\) and \(b\) represent the output and bias of the layer, respectively. By integrating both low-rank and sparse components, our method outperforms pruning-only approach, enhancing feature extraction and achieving higher accuracy after fine-tuning.

\begin{table*}[thb]
    \centering
    \renewcommand{\arraystretch}{1}
\resizebox{\linewidth}{!}{
\begin{tabular}{cccccccccc}
\hline
Model                         & Method                                & PIQA                                                          & BoolQ                                                         & HellaS                                                        & Wino                                                          & ARC-e                                                         & ARC-c                                                         & Ave                                                           & $\Delta$                                                      \\ \hline
                              & Dense                                 & 78.07                                                         & 77.71                                                         & 57.14                                                         & 68.90                                                         & 76.35                                                         & 43.60                                                         & 66.96                                                         & -                                                             \\ \cline{2-10} 
                              & SparseGPT*                            & 76.09                                                         & 76.94                                                         & 55.63                                                         & \textbf{68.35}                                                & 73.32                                                         & 41.04                                                         & 65.22                                                         & -1.74                                                         \\
                              & Wanda*                                & 77.69                                                         & 76.82                                                         & 54.57                                                         & 67.75                                                         & 74.28                                                         & 41.21                                                         & 65.39                                                         & -1.57                                                         \\
\multirow{-4}{*}{LLaMA2-7B}   & \cellcolor[HTML]{EFEFEF}\textbf{Ours}                                  & \cellcolor[HTML]{EFEFEF}\textbf{78.18}                        & \cellcolor[HTML]{EFEFEF}\textbf{77.03}                        & \cellcolor[HTML]{EFEFEF}\textbf{57.09}                        & \cellcolor[HTML]{EFEFEF}67.72                                 & \cellcolor[HTML]{EFEFEF}\textbf{75.17}                        & \cellcolor[HTML]{EFEFEF}\textbf{43.26}                        & \cellcolor[HTML]{EFEFEF}\textbf{66.41}                        & \cellcolor[HTML]{EFEFEF}\textbf{-0.55}                        \\ \hline
                              & Dense                                 & 80.14                                                         & 82.08                                                         & 60.02                                                         & 73.64                                                         & 81.40                                                         & 51.19                                                         & 71.41                                                         & -                                                             \\ \cline{2-10} 
                              & SparseGPT*                            & 78.51                                                         & \textbf{81.91}                                                         & 57.40                                                         & 71.82                                                         & 79.22                                                         & 48.14                                                         & 69.50                                                         & -1.91                                                         \\
                              & Wanda*                                & 78.18                                                         & 78.75                                                         & 56.95                                                         & 72.22                                                         & 79.01                                                         & 48.82                                                         & 68.99                                                         & -2.42                                                         \\
\multirow{-4}{*}{LLaMA3-8B}   & \cellcolor[HTML]{EFEFEF}\textbf{Ours} & \cellcolor[HTML]{EFEFEF}{\color[HTML]{000000} \textbf{79.32}} & \cellcolor[HTML]{EFEFEF}{\color[HTML]{000000} 80.75} & \cellcolor[HTML]{EFEFEF}{\color[HTML]{000000} \textbf{58.67}} & \cellcolor[HTML]{EFEFEF}{\color[HTML]{000000} \textbf{72.48}} & \cellcolor[HTML]{EFEFEF}{\color[HTML]{000000} \textbf{80.60}} & \cellcolor[HTML]{EFEFEF}{\color[HTML]{000000} \textbf{50.68}} & \cellcolor[HTML]{EFEFEF}{\color[HTML]{000000} \textbf{70.42}} & \cellcolor[HTML]{EFEFEF}{\color[HTML]{000000} \textbf{-0.99}} \\ \hline
                              & Dense                                 & 78.51                                                         & 84.52                                                         & 72.77                                                         & 60.01                                                         & 80.56                                                         & 48.63                                                         & 70.83                                                         & -                                                             \\ \cline{2-10} 
                              & SparseGPT*                            & 79.03                                                         & 84.54                                                         & 71.69                                                         & 57.13                                                         & 80.44                                                         & 51.21                                                         & 70.67                                                         & -0.16                                                         \\
                              & Wanda*                                & \textbf{79.11}                                                & 84.71                                                         & 70.17                                                         & 56.64                                                         & 79.80                                                         & 50.09                                                         & 70.09                                                         & -0.74                                                         \\
\multirow{-4}{*}{Qwen2.5-7B}  & \cellcolor[HTML]{EFEFEF}\textbf{Ours} & \cellcolor[HTML]{EFEFEF}78.84                                 & \cellcolor[HTML]{EFEFEF}\textbf{85.44}                        & \cellcolor[HTML]{EFEFEF}\textbf{72.06}                        & \cellcolor[HTML]{EFEFEF}\textbf{58.20}                        & \cellcolor[HTML]{EFEFEF}\textbf{81.82}                        & \cellcolor[HTML]{EFEFEF}\textbf{52.64}                        & \cellcolor[HTML]{EFEFEF}\textbf{71.50}                        & \cellcolor[HTML]{EFEFEF}\textbf{+0.67}                        \\ \hline
                              & Dense                                 & 81.12                                                         & 85.54                                                         & 75.37                                                         & 63.39                                                         & 82.37                                                         & 55.80                                                         & 73.93                                                         & -                                                             \\ \cline{2-10} 
                              & SparseGPT*                            & 80.45                                                         & 87.63                                                         & 73.52                                                         & 60.78                                                         & 82.42                                                         & 55.03                                                         & 73.31                                                         & -0.62                                                         \\
                              & Wanda*                                & 79.71                                                         & 87.70                                                         & 73.48                                                         & 60.44                                                         & 82.62                                                         & 54.78                                                         & 73.12                                                         & -0.81                                                         \\
\multirow{-4}{*}{Qwen2.5-14B} & \cellcolor[HTML]{EFEFEF}\textbf{Ours} & \cellcolor[HTML]{EFEFEF}\textbf{81.39}                        & \cellcolor[HTML]{EFEFEF}\textbf{87.74}                        & \cellcolor[HTML]{EFEFEF}\textbf{74.03}                        & \cellcolor[HTML]{EFEFEF}\textbf{61.58}                        & \cellcolor[HTML]{EFEFEF}\textbf{84.34}                        & \cellcolor[HTML]{EFEFEF}\textbf{56.06}                        & \cellcolor[HTML]{EFEFEF}\textbf{74.19}                        & \cellcolor[HTML]{EFEFEF}\textbf{+0.26}                        \\ \hline
\end{tabular}}
    \caption{Zero-shot tasks accuracy (\%) of LLaMA and Qwen2.5 models at 50\% compression rate after fine-tuning with different pruning methods. * indicates models with
LoRA fine-tuning, which introduces an additional parameter.} 
    \label{tab:finetuning}
    \vspace{-1mm}
\end{table*}

\section{Evaluation}\label{sec:exp}

A comprehensive evaluation of the LLaMA and Qwen2.5 model family has been conducted to assess the effectiveness of SSLC. Detailed experimental setups, pre-trained models, datasets, and baselines are provided in Appendix~\ref{sec:appendix:set}. Here, we present the performance analysis of the compressed models, focusing on perplexity and zero-shot capability. Additionally, we performed ablation studies to illustrate the impact of key hyperparameters such as rank, iteration count and weight preservation strategy. Finally, we evaluated the acceleration potential of our method using the simulated ViTCOD~\cite{you2023vitcod} accelerator, as detailed in Appendix~\ref{sec::vitcod}.

\subsection{Compression Rate Efficiency Comparison}\label{sec:compresion_rate_spread}

\begin{figure}[t]
    \centering
    \includegraphics[width=0.95\linewidth]{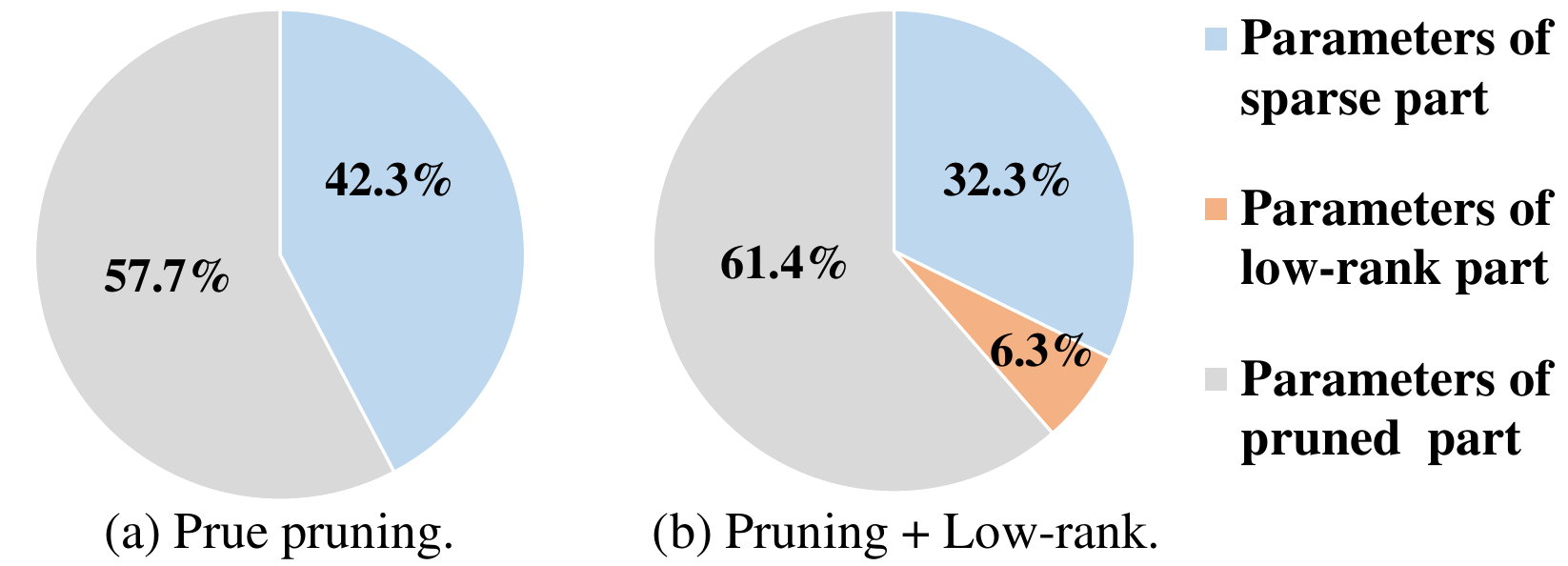}
    \caption{Retaining 80\% of the total salience, the pure pruning method necessitates keeping the top 42.3\% of parameters, which compresses 57.7\% parameters. In contrast, the synergistic method requires only the top 32.3\% of parameters to form a sparse matrix, and with the additional 6.25\% from the low-rank matrix. The overall reserved parameter ratio (38.6\%) remains lower than that of the pure pruning method (42.3\%), which shows the compression ``rate spread'' of 3.7\%.}
    \label{fig:parameter_distribution_pie}
\end{figure}

As quantified in Figure~\ref{fig:parameter_distribution_pie}, when retaining 80\% of the original weight salience (as measured by Eq.~\ref{eq:prune}), our synergistic method requires only 38.6\% parameter retention. This represents a 3.7\% absolute reduction compared to the pure pruning baseline (42.3\%). The efficiency gain originates from decoupling parameters into complementary components: a 32.3\% sparse matrix preserves the most crucial full-rank components for knowledge retention, while an additional 6.25\% from the low-rank approximation encodes the essential structure.

\subsection{Language Modeling and Zero-shot Tasks}\label{sec:exp:lang_model}

Table~\ref{tab:language_modelling} shows the performance of sparse LLM models at a uniform sparsity rate of 50\%. Our method, SSLC, achieves state-of-the-art results across both language modeling and zero-shot tasks, significantly outperforming baselines such as Wanda and DSnoT on various datasets, including C4 and WikiText-2. Moreover, our experiments demonstrate that the compressed models such as Qwen2.5-14B with SSLC (approximately 7B effective parameters) outperforms the native dense Qwen2.5-7B on zero-shot tasks, achieving an average improvement of 1.1\% on benchmarks. These results highlighting that sparsity-based compression not only reduces parameter counts but better preserves the original models's capabilities compared to architecturally constrained smaller models.

\subsection{Fine-tuning Sparse LLMs}\label{sec:exp:0_shot}To bridge the remaining performance gap, we further explore parameter-efficient fine-tuning strategies. As shown in Figure~\ref{fig:fine-turning}, unlike other methods such as Wanda and SparseGPT, which introduce additional parameters during adaptation, SSLC leverages its low-rank structure for parameter-efficient fine-tuning. As detailed in Table~\ref{tab:finetuning}, after fine-tuning on alpaca datasets, SSLC not only surpasses Wanda and SparseGPT with LoRA but also nearly recovers the full accuracy of the original dense model, particularly on LLaMA2-7B and Qwen 2.5 models. This demonstrates that SSLC enables sparse LLMs to retain high performance under tight parameter budgets, making it especially suitable for practical deployment scenarios where storage and efficiency are critical.

\begin{figure}[ht]
    \centering 
  \includegraphics[width=0.78\linewidth]{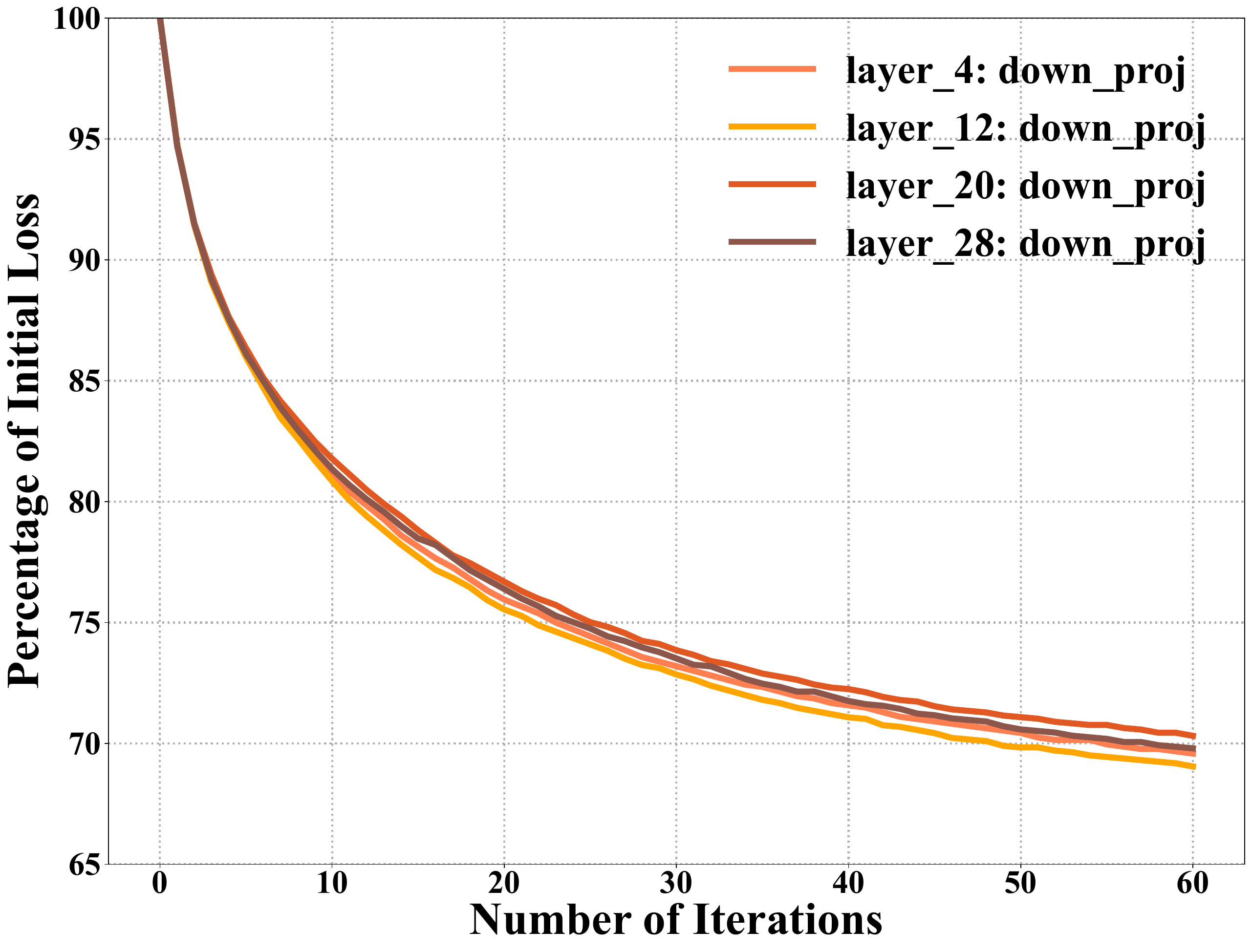}
\caption{ The current decomposition loss, denoted as \(\left \| (W-L_t-S_t)X\right \| _F\), for the down projection matrices of different layers in LLaMA2-7B varies as a percentage of the initial loss with respect to the number of iterations.}
\label{fig:iterative_loss}
\end{figure}

\subsection{Ablation Study}\label{sec:ablation study}
We conduct ablation studies to assess the contribution of key hyperparameters in our SSLC method. As shown in Figure \ref{fig:iterative_loss}, the reconstruction error decreases rapidly across network layers when $T$ increases from 0 to 20, and notably stabilizes after 40 iterations, indicating robust convergence behavior of our method. Our experiments on C4 and WikiText-2 datasets (Table ~\ref{tab:ablate_number_of_iterations}) further confirm that the model achieves stable performance after 40 iterations, with optimal results appearing at $T$=60. After balancing computational efficiency with performance requirements, we ultimately selected 40 iterations as the experimental setting. This choice maintains model effectiveness while significantly reducing computational overhead (40 iterations consume 33\% less resources than 60 iterations).

\begin{table}[thb]
    \centering 
    \renewcommand{\arraystretch}{0.7}
    \begin{tabular}{cccc}
    \toprule
    Iteration & Wikitext-2 & C4    & Average\\ \midrule
    0      & 7.35      & 9.75  & 8.55\\
    10     & 6.84      & 9.16  & 8.00\\
    20     & 6.74      & 8.99  & 7.87\\
    30     & 6.67      & 8.91  & 7.79\\
    40     & 6.61      & 8.87  & 7.74\\
    50     & 6.59      & 8.85  & 7.72\\
    \cellcolor[HTML]{EFEFEF}\textbf{60}     & \cellcolor[HTML]{EFEFEF}\textbf{6.58}  & \cellcolor[HTML]{EFEFEF}\textbf{8.83} & \cellcolor[HTML]{EFEFEF}\textbf{7.71}\\ \bottomrule
    \end{tabular}
    \caption{Perplexity for LLaMA2-7B with 50\% parameters remaining at different numbers of iterations.} 
    \label{tab:ablate_number_of_iterations}
    \vspace{-1mm}
\end{table}

\begin{figure}[h]
    \centering 
  \includegraphics[width=0.85\linewidth]{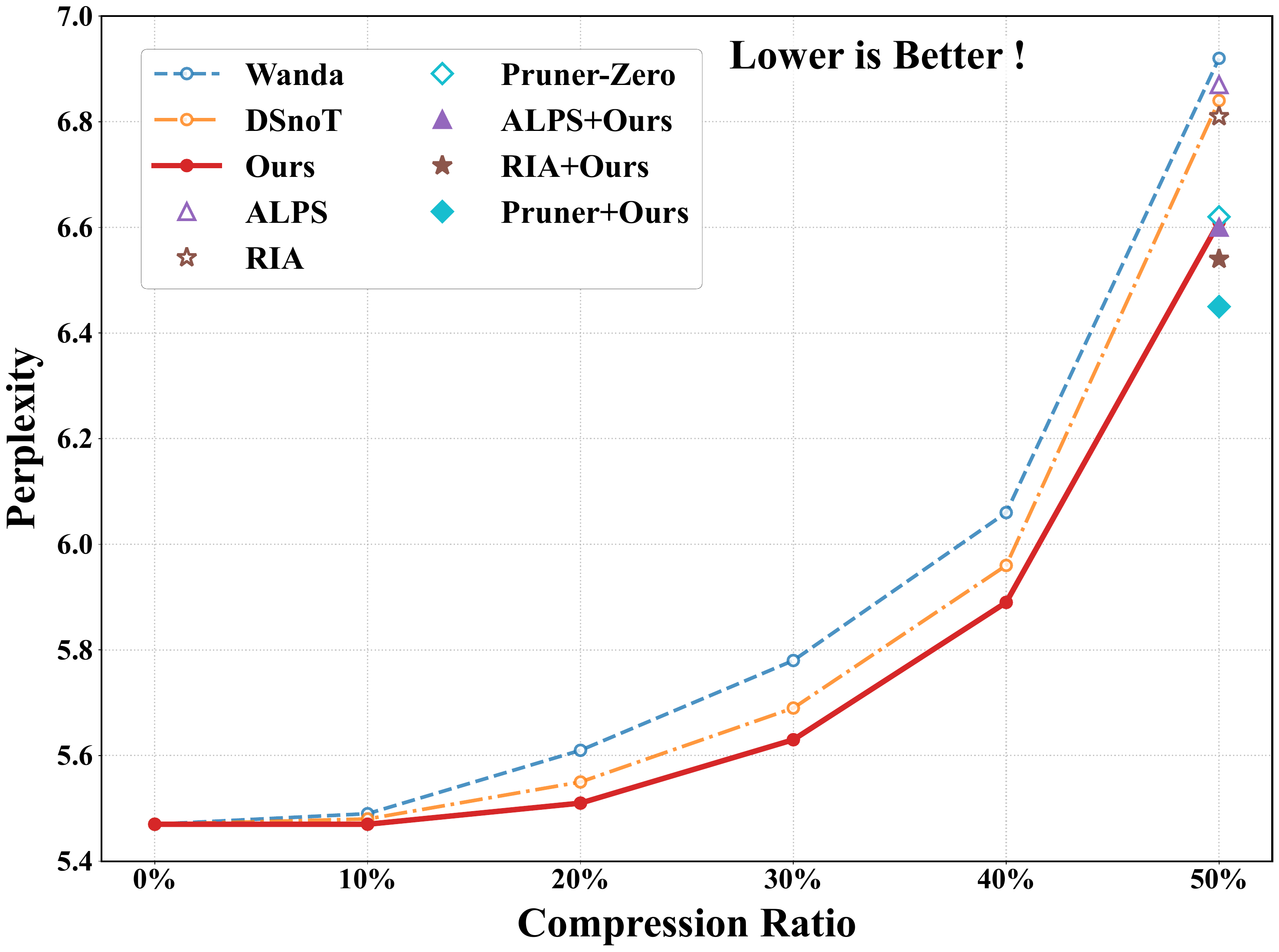}
\caption{Performance of LLaMA2-7B on the WikiText-2 dataset under varying pruning ratios. Hollow markers denote standalone pruning methods, while solid markers represent our synergistic compression approach.}
\label{fig:pruning_ratio}
\end{figure}

To rigorously validate the effectiveness of our SSLC framework, we performed systematic evaluations across various sparsity configurations. As evidenced by the experimental results presented in Figure~\ref{fig:pruning_ratio}, our method demonstrates consistent superiority over baseline approaches under varying pruning intensities, ranging from 10\% to 50\% sparsity levels. The performance gap becomes particularly pronounced at higher sparsity rates, highlighting the efficiency of our approach in preserving model capabilities even under aggressive compression. Furthermore, by integrating our SSLC framework with existing pruning techniques, the enhanced approaches achieve significantly better performance than their vanilla implementations. 

For detailed ablation studies on the other three key hyperparameters: (1)~the number of retained ranks, (2)~the salience-based weight preservation strategy, and (3)~random seed initialization, alongside a comparative analysis of pruning methods under the SSLC framework, refer to Appendix ~\ref{sec:method:ablation}.

\subsection{Acceleration Performance}\label{sec:method:speedup_sim}
To evaluate the acceleration of unstructured pruning, we employ the ViTCoD accelerator simulator to assess SSLC at a 50\% compression ratio. As detailed in Table~\ref{tab:speed}, our method achieves speedups of \textbf{1.74$\times$} (MHA) and \textbf{1.84$\times$} (FFN) for LLaMA2-7B, and \textbf{1.63$\times$} (MHA) and \textbf{1.85$\times$} (FFN) for Qwen2.5-7B.

\begin{table}[h]
    \centering
    \setlength{\tabcolsep}{4.5pt}
    \resizebox{\linewidth}{!}{
    \begin{tabular}{ccccc}
    \hline
    Model    & \multicolumn{2}{c}{LLaMA2-7B} & \multicolumn{2}{c}{Qwen2.5-7B} \\ \hline
    Module   & MHA        & FFN         & MHA         & FFN         \\
    Dense    & 16384      & 33024       & 7168        & 49728       \\ \hline
    Sparse   & 8364.2     & 16535.3     & 3705.7      & 24764.5     \\
    Low-rank & 1024       & 1416        & 704         & 2112        \\
    Sum      & 9388.2     & 17951.3     & 4409.7      & 26876.5     \\ \hline
    Speedup  & 1.74$\times$ & 1.84$\times$ & 1.63$\times$ & 1.85$\times$ \\ \hline
    \end{tabular}}
    \caption{Runtime (cycles) and speedup across modules in LLaMA2-7B and Qwen2.5-7B. "Cycles" denotes computational cycles required by the ViTCoD accelerator.} 
    \label{tab:speed}
    \vspace{-1mm}
\end{table}

\begin{table}[h]
    \centering
    \label{tab:realtime}
    \renewcommand{\arraystretch}{0.8}
    \begin{tabular}{ccccc}
    \toprule
    Model & Dense & 50\% & 60\% & 70\%  \\
    \midrule
    LLaMA2-7B  & 53.79 & 72.12 & 77.87 & \textbf{89.87}  \\
    LLaMA1-7B  & 54.07 & 73.02 & 79.14 & \textbf{91.25}  \\
    \bottomrule
    \end{tabular}
    \caption{Real-world throughput (tokens/sec) at varying sparsity levels}
\end{table}

For real-world memory-bound inference, we evaluate SSLC across sparsity levels from 50\% to 70\% using nm-vLLM~\cite{neuralmagic_nm_vllm_2024}. With 1024-token generation over 5 prompts, SSLC achieves throughput speedups of \textbf{1.34$\times$--1.69$\times$} in bandwidth bottleneck.

\section{Conclusion}\label{sec:conclusion}

In this paper, we systematically analyze the strengths and weaknesses of two previously independent compression techniques for LLMs: pruning and low-rank approximation. Based on the theoretical analysis, SSLC (Synergistic Sparse and Low-Rank Compression) is introduced for efficient LLM deployment, which maximizes the energy in the low-rank component using orthogonal bases, while simultaneously achieving discrete full-rank information in the sparse part. By modeling the joint compression for LLMs as a unified optimization problem, we apply an iterative optimization algorithm that offers a novel theoretical perspective and achieves significant performance improvements in practice. Experiments on language modeling and zero-shot tasks show that our method significantly outperforms previous compression approaches. Furthermore, comprehensive fine-tuning experiments demonstrate SSLC's effectiveness in restoring model accuracy, validating its practicality for real-world deployment.

\section*{Limitations}

Our proposed synergistic sparse and low-rank compression method is formulated as an iterative optimization problem. While this approach necessitates additional computation during the pruning phase, we have strategically optimized the algorithm to minimize both time and memory consumption. As a result, the pruning process completes in approximately 30 minutes for 7B models and about 1 hour for 14B models on standard hardware configurations. Despite these efficiency gains, our method currently applies uniform compression ratios across all Transformer layers, which may not fully exploit the varying sensitivities of different layers. Future work will focus on exploring theoretically grounded metrics for assessing layer criticality—potentially through gradient-weighted Hessian analysis—to enable dynamic, layer-wise compression policies that achieves Pareto-efficient trade-offs between accuracy and computational cost.

\bibliography{acl_latex}

\begin{thebibliography}{53}
\expandafter\ifx\csname natexlab\endcsname\relax\def\natexlab#1{#1}\fi

\bibitem[{Achiam et~al.(2023)Achiam, Adler, Agarwal, Ahmad, Akkaya, Aleman, Almeida, Altenschmidt, Altman, Anadkat et~al.}]{achiam2023gpt}
Josh Achiam, Steven Adler, Sandhini Agarwal, Lama Ahmad, Ilge Akkaya, Florencia~Leoni Aleman, Diogo Almeida, Janko Altenschmidt, Sam Altman, Shyamal Anadkat, et~al. 2023.
\newblock Gpt-4 technical report.
\newblock \emph{arXiv preprint arXiv:2303.08774}.

\bibitem[{Aghajanyan et~al.(2020)Aghajanyan, Zettlemoyer, and Gupta}]{aghajanyan2020intrinsic}
Armen Aghajanyan, Luke Zettlemoyer, and Sonal Gupta. 2020.
\newblock Intrinsic dimensionality explains the effectiveness of language model fine-tuning.
\newblock \emph{arXiv preprint arXiv:2012.13255}.

\bibitem[{Bisk et~al.(2020)Bisk, Zellers, Bras, Gao, and Choi}]{Bisk2020piqa}
Yonatan Bisk, Rowan Zellers, Ronan~Le Bras, Jianfeng Gao, and Yejin Choi. 2020.
\newblock Piqa: Reasoning about physical commonsense in natural language.
\newblock In \emph{Thirty-Fourth AAAI Conference on Artificial Intelligence}.

\bibitem[{Bubeck et~al.(2023)Bubeck, Chandrasekaran, Eldan, Gehrke, Horvitz, Kamar, Lee, Lee, Li, Lundberg, Nori, Palangi, Ribeiro, and Zhang}]{bubeck2023sparks}
Sébastien Bubeck, Varun Chandrasekaran, Ronen Eldan, Johannes Gehrke, Eric Horvitz, Ece Kamar, Peter Lee, Yin~Tat Lee, Yuanzhi Li, Scott Lundberg, Harsha Nori, Hamid Palangi, Marco~Tulio Ribeiro, and Yi~Zhang. 2023.
\newblock \href {http://arxiv.org/abs/2303.12712} {Sparks of artificial general intelligence: Early experiments with gpt-4}.

\bibitem[{Chen et~al.(2023)Chen, Ding, Yadav, Zharkov, and Liang}]{chen2023lorashear}
Tianyi Chen, Tianyu Ding, Badal Yadav, Ilya Zharkov, and Luming Liang. 2023.
\newblock Lorashear: Efficient large language model structured pruning and knowledge recovery.
\newblock \emph{arXiv preprint arXiv:2310.18356}.

\bibitem[{Clark et~al.(2019)Clark, Lee, Chang, Kwiatkowski, Collins, and Toutanova}]{clark-etal-2019-boolq}
Christopher Clark, Kenton Lee, Ming-Wei Chang, Tom Kwiatkowski, Michael Collins, and Kristina Toutanova. 2019.
\newblock \href {https://doi.org/10.18653/v1/N19-1300} {{B}ool{Q}: Exploring the surprising difficulty of natural yes/no questions}.
\newblock In \emph{Proceedings of the 2019 Conference of the North {A}merican Chapter of the Association for Computational Linguistics: Human Language Technologies, Volume 1 (Long and Short Papers)}, pages 2924--2936, Minneapolis, Minnesota. Association for Computational Linguistics.

\bibitem[{Clark et~al.(2018)Clark, Cowhey, Etzioni, Khot, Sabharwal, Schoenick, and Tafjord}]{allenai:arc}
Peter Clark, Isaac Cowhey, Oren Etzioni, Tushar Khot, Ashish Sabharwal, Carissa Schoenick, and Oyvind Tafjord. 2018.
\newblock Think you have solved question answering? try arc, the ai2 reasoning challenge.
\newblock \emph{arXiv:1803.05457v1}.

\bibitem[{Dettmers et~al.(2022)Dettmers, Lewis, Belkada, and Zettlemoyer}]{dettmers2022llmint8}
Tim Dettmers, Mike Lewis, Younes Belkada, and Luke Zettlemoyer. 2022.
\newblock {LLM}.int8(): 8-bit matrix multiplication for transformers at scale.
\newblock In \emph{Advances in Neural Information Processing Systems}.

\bibitem[{Dettmers et~al.(2023)Dettmers, Svirschevski, Egiazarian, Kuznedelev, Frantar, Ashkboos, Borzunov, Hoefler, and Alistarh}]{dettmers2023spqr}
Tim Dettmers, Ruslan Svirschevski, Vage Egiazarian, Denis Kuznedelev, Elias Frantar, Saleh Ashkboos, Alexander Borzunov, Torsten Hoefler, and Dan Alistarh. 2023.
\newblock \href {http://arxiv.org/abs/2306.03078} {Spqr: A sparse-quantized representation for near-lossless llm weight compression}.

\bibitem[{Dong et~al.(2024)Dong, Li, Tang, Liu, Pan, Wang, and Chu}]{dong2024pruner}
Peijie Dong, Lujun Li, Zhenheng Tang, Xiang Liu, Xinglin Pan, Qiang Wang, and Xiaowen Chu. 2024.
\newblock Pruner-zero: Evolving symbolic pruning metric from scratch for large language models.
\newblock \emph{arXiv preprint arXiv:2406.02924}.

\bibitem[{Frantar and Alistarh(2022)}]{frantar2022optimal}
Elias Frantar and Dan Alistarh. 2022.
\newblock Optimal brain compression: A framework for accurate post-training quantization and pruning.
\newblock \emph{Advances in Neural Information Processing Systems}, 35:4475--4488.

\bibitem[{Frantar and Alistarh(2023)}]{frantar2023sparsegpt}
Elias Frantar and Dan Alistarh. 2023.
\newblock \href {http://arxiv.org/abs/2301.00774} {{SparseGPT}: Massive language models can be accurately pruned in one-shot}.

\bibitem[{Frantar et~al.(2023)Frantar, Ashkboos, Hoefler, and Alistarh}]{frantar2023gptq}
Elias Frantar, Saleh Ashkboos, Torsten Hoefler, and Dan Alistarh. 2023.
\newblock {GPTQ}: Accurate post-training compression for generative pretrained transformers.
\newblock In \emph{International Conference on Learning Representations}.

\bibitem[{Gao et~al.(2021)Gao, Tow, Biderman, Black, DiPofi, Foster, Golding, Hsu, McDonell, Muennighoff et~al.}]{eval-harness}
Leo Gao, Jonathan Tow, Stella Biderman, Sid Black, Anthony DiPofi, Charles Foster, Laurence Golding, Jeffrey Hsu, Kyle McDonell, Niklas Muennighoff, et~al. 2021.
\newblock A framework for few-shot language model evaluation.
\newblock \emph{Version v0. 0.1. Sept}.

\bibitem[{Grattafiori et~al.(2024)Grattafiori, Dubey, Jauhri, Pandey, Kadian, Al-Dahle, Letman, Mathur, Schelten, Vaughan et~al.}]{grattafiori2024llama3herdmodels}
Aaron Grattafiori, Abhimanyu Dubey, Abhinav Jauhri, Abhinav Pandey, Abhishek Kadian, Ahmad Al-Dahle, Aiesha Letman, Akhil Mathur, Alan Schelten, Alex Vaughan, et~al. 2024.
\newblock \href {http://arxiv.org/abs/2407.21783} {The llama 3 herd of models}.

\bibitem[{Hassibi and Stork(1992)}]{hassibi1992second}
Babak Hassibi and David Stork. 1992.
\newblock Second order derivatives for network pruning: Optimal brain surgeon.
\newblock \emph{Advances in neural information processing systems}, 5.

\bibitem[{Hassibi et~al.(1993)Hassibi, Stork, and Wolff}]{hassibi1993obs}
Babak Hassibi, David~G Stork, and Gregory~J Wolff. 1993.
\newblock Optimal brain surgeon and general network pruning.
\newblock In \emph{IEEE International Conference on Neural Networks}.

\bibitem[{Hsu et~al.(2022)Hsu, Hua, Chang, Lou, Shen, and Jin}]{hsu2022language}
Yen-Chang Hsu, Ting Hua, Sungen Chang, Qian Lou, Yilin Shen, and Hongxia Jin. 2022.
\newblock Language model compression with weighted low-rank factorization.
\newblock \emph{arXiv preprint arXiv:2207.00112}.

\bibitem[{Hu et~al.(2021)Hu, Shen, Wallis, Allen-Zhu, Li, Wang, Wang, and Chen}]{hu2021lora}
Edward~J Hu, Yelong Shen, Phillip Wallis, Zeyuan Allen-Zhu, Yuanzhi Li, Shean Wang, Lu~Wang, and Weizhu Chen. 2021.
\newblock \href {http://arxiv.org/abs/2106.09685} {Lora: Low-rank adaptation of large language models}.

\bibitem[{Huang et~al.(2024)Huang, Qin, Liu, Li, Liu, Benini, Magno, and Qi}]{huang2024slim}
Wei Huang, Haotong Qin, Yangdong Liu, Yawei Li, Xianglong Liu, Luca Benini, Michele Magno, and Xiaojuan Qi. 2024.
\newblock Slim-llm: Salience-driven mixed-precision quantization for large language models.
\newblock \emph{arXiv preprint arXiv:2405.14917}.

\bibitem[{Huang et~al.(2025)Huang, Zhang, Zheng, Liu, Lin, Yao, and Ji}]{huang2025dynamiclowranksparseadaptation}
Weizhong Huang, Yuxin Zhang, Xiawu Zheng, Yang Liu, Jing Lin, Yiwu Yao, and Rongrong Ji. 2025.
\newblock \href {http://arxiv.org/abs/2502.14816} {Dynamic low-rank sparse adaptation for large language models}.

\bibitem[{LeCun et~al.(1989)LeCun, Denker, and Solla}]{lecun1990obd}
Yann LeCun, John~S Denker, and Sara~A Solla. 1989.
\newblock Optimal brain damage.
\newblock In \emph{Advances in Neural Information Processing Systems}.

\bibitem[{Li et~al.(2024)Li, Tang, and Zhang}]{li2024lorap}
Guangyan Li, Yongqiang Tang, and Wensheng Zhang. 2024.
\newblock Lorap: Transformer sub-layers deserve differentiated structured compression for large language models.
\newblock \emph{arXiv preprint arXiv:2404.09695}.

\bibitem[{Li et~al.(2023{\natexlab{a}})Li, Yu, Liang, He, Karampatziakis, Chen, and Zhao}]{li2023loftq}
Yixiao Li, Yifan Yu, Chen Liang, Pengcheng He, Nikos Karampatziakis, Weizhu Chen, and Tuo Zhao. 2023{\natexlab{a}}.
\newblock Loftq: Lora-fine-tuning-aware quantization for large language models.
\newblock \emph{arXiv preprint arXiv:2310.08659}.

\bibitem[{Li et~al.(2023{\natexlab{b}})Li, Yu, Zhang, Liang, He, Chen, and Zhao}]{li2023losparse}
Yixiao Li, Yifan Yu, Qingru Zhang, Chen Liang, Pengcheng He, Weizhu Chen, and Tuo Zhao. 2023{\natexlab{b}}.
\newblock Losparse: Structured compression of large language models based on low-rank and sparse approximation.
\newblock In \emph{International Conference on Machine Learning}, pages 20336--20350. PMLR.

\bibitem[{Liu et~al.(2025)Liu, Zhao, Fedorov, Soran, Choudhary, Krishnamoorthi, Chandra, Tian, and Blankevoort}]{liu2025spinquantllmquantizationlearned}
Zechun Liu, Changsheng Zhao, Igor Fedorov, Bilge Soran, Dhruv Choudhary, Raghuraman Krishnamoorthi, Vikas Chandra, Yuandong Tian, and Tijmen Blankevoort. 2025.
\newblock \href {http://arxiv.org/abs/2405.16406} {Spinquant: Llm quantization with learned rotations}.

\bibitem[{Ma et~al.(2023)Ma, Fang, and Wang}]{ma2023llmv3}
Xinyin Ma, Gongfan Fang, and Xinchao Wang. 2023.
\newblock \href {http://arxiv.org/abs/2305.11627} {Llm-pruner: On the structural pruning of large language models}.
\newblock Version 3.

\bibitem[{Meng et~al.(2024)Meng, Behdin, Wang, and Mazumder}]{meng2024alpsimprovedoptimizationhighly}
Xiang Meng, Kayhan Behdin, Haoyue Wang, and Rahul Mazumder. 2024.
\newblock \href {http://arxiv.org/abs/2406.07831} {Alps: Improved optimization for highly sparse one-shot pruning for large language models}.

\bibitem[{Merity et~al.(2016)Merity, Xiong, Bradbury, and Socher}]{merity2016wiki2}
Stephen Merity, Caiming Xiong, James Bradbury, and Richard Socher. 2016.
\newblock Pointer sentinel mixture models.
\newblock \emph{arXiv preprint arXiv:1609.07843}.

\bibitem[{NeuralMagic(2024)}]{neuralmagic_nm_vllm_2024}
NeuralMagic. 2024.
\newblock \href {https://github.com/neuralmagic/nm-vllm} {nm-vllm: Neuralmagic's inference engine for {vLLM}}.
\newblock \url{https://github.com/neuralmagic/nm-vllm}.
\newblock Accessed: 2025-09-01.

\bibitem[{Raffel et~al.(2020)Raffel, Shazeer, Roberts, Lee, Narang, Matena, Zhou, Li, and Liu}]{raffel2020c4}
Colin Raffel, Noam Shazeer, Adam Roberts, Katherine Lee, Sharan Narang, Michael Matena, Yanqi Zhou, Wei Li, and Peter~J Liu. 2020.
\newblock Exploring the limits of transfer learning with a unified text-to-text transformer.
\newblock \emph{Journal of machine learning research}, 21(140):1--67.

\bibitem[{Saha et~al.(2024)Saha, Sagan, Srivastava, Goldsmith, and Pilanci}]{saha2024compressinglargelanguagemodels}
Rajarshi Saha, Naomi Sagan, Varun Srivastava, Andrea~J. Goldsmith, and Mert Pilanci. 2024.
\newblock \href {http://arxiv.org/abs/2405.18886} {Compressing large language models using low rank and low precision decomposition}.

\bibitem[{Sakaguchi et~al.(2019)Sakaguchi, Bras, Bhagavatula, and Choi}]{ai2:winogrande}
Keisuke Sakaguchi, Ronan~Le Bras, Chandra Bhagavatula, and Yejin Choi. 2019.
\newblock \href {http://arxiv.org/abs/1907.10641} {Winogrande: An adversarial winograd schema challenge at scale}.

\bibitem[{Scao et~al.(2022)Scao, Fan, Akiki, Pavlick, Ilić, Hesslow, Castagné, Luccioni, Yvon, Gallé et~al.}]{scao2022bloom}
Teven~Le Scao, Angela Fan, Christopher Akiki, Ellie Pavlick, Suzana Ilić, Daniel Hesslow, Roman Castagné, Alexandra~Sasha Luccioni, François Yvon, Matthias Gallé, et~al. 2022.
\newblock \href {http://arxiv.org/abs/2211.05100} {Bloom: A 176b-parameter open-access multilingual language model}.

\bibitem[{Sun et~al.(2023)Sun, Liu, Bair, and Kolter}]{sun2023simple}
Mingjie Sun, Zhuang Liu, Anna Bair, and Zico Kolter. 2023.
\newblock \href {http://arxiv.org/abs/2306.11695} {A simple and effective pruning approach for large language models}.

\bibitem[{Taori et~al.(2023)Taori, Gulrajani, Zhang, Dubois, Li, Guestrin, Liang, and Hashimoto}]{taori2023alpaca}
Rohan Taori, Ishaan Gulrajani, Tianyi Zhang, Yann Dubois, Xuechen Li, Carlos Guestrin, Percy Liang, and Tatsunori~B. Hashimoto. 2023.
\newblock Stanford alpaca: An instruction-following llama model.
\newblock \url{https://github.com/tatsu-lab/stanford_alpaca}.
\newblock Accessed: 2023-08-09.

\bibitem[{Touvron et~al.(2023{\natexlab{a}})Touvron, Lavril, Izacard, Martinet, Lachaux, Lacroix, Rozi{\`e}re, Goyal, Hambro, Azhar, Rodriguez, Joulin, Grave, and Lample}]{touvron2023llama}
Hugo Touvron, Thibaut Lavril, Gautier Izacard, Xavier Martinet, Marie-Anne Lachaux, Timoth{\'e}e Lacroix, Baptiste Rozi{\`e}re, Naman Goyal, Eric Hambro, Faisal Azhar, Aurelien Rodriguez, Armand Joulin, Edouard Grave, and Guillaume Lample. 2023{\natexlab{a}}.
\newblock \href {http://arxiv.org/abs/2302.13971} {{LLaMA}: Open and efficient foundation language models}.

\bibitem[{Touvron et~al.(2023{\natexlab{b}})Touvron, Martin, Stone, Albert, Almahairi, Babaei, Bashlykov, Batra, Bhargava, Bhosale et~al.}]{touvron2023llama2}
Hugo Touvron, Louis Martin, Kevin Stone, Peter Albert, Amjad Almahairi, Yasmine Babaei, Nikolay Bashlykov, Soumya Batra, Prajjwal Bhargava, Shruti Bhosale, et~al. 2023{\natexlab{b}}.
\newblock Llama 2: Open foundation and fine-tuned chat models.
\newblock \emph{arXiv preprint arXiv:2307.09288}.

\bibitem[{Wang et~al.(2024)Wang, Zheng, Wan, and Zhang}]{wang2024svd}
Xin Wang, Yu~Zheng, Zhongwei Wan, and Mi~Zhang. 2024.
\newblock Svd-llm: Truncation-aware singular value decomposition for large language model compression.
\newblock \emph{arXiv preprint arXiv:2403.07378}.

\bibitem[{Wei et~al.(2022)Wei, Tay, Bommasani, Raffel, Zoph, Borgeaud, Yogatama, Bosma, Zhou, Metzler, Chi, Hashimoto, Vinyals, Liang, Dean, and Fedus}]{wei2022emergent}
Jason Wei, Yi~Tay, Rishi Bommasani, Colin Raffel, Barret Zoph, Sebastian Borgeaud, Dani Yogatama, Maarten Bosma, Denny Zhou, Donald Metzler, Ed~H. Chi, Tatsunori Hashimoto, Oriol Vinyals, Percy Liang, Jeff Dean, and William Fedus. 2022.
\newblock Emergent abilities of large language models.
\newblock In \emph{Transactions on Machine Learning Research}.

\bibitem[{Wright et~al.(2009)Wright, Ganesh, Rao, Peng, and Ma}]{wright2009robust}
John Wright, Arvind Ganesh, Shankar Rao, Yigang Peng, and Yi~Ma. 2009.
\newblock Robust principal component analysis: Exact recovery of corrupted low-rank matrices via convex optimization.
\newblock \emph{Advances in neural information processing systems}, 22.

\bibitem[{Xiang et~al.(2012)Xiang, Zhu, Shen, and Ye}]{xiang2012optimal}
Shuo Xiang, Yunzhang Zhu, Xiaotong Shen, and Jieping Ye. 2012.
\newblock Optimal exact least squares rank minimization.
\newblock In \emph{Proceedings of the 18th ACM SIGKDD international conference on Knowledge discovery and data mining}, pages 480--488.

\bibitem[{Xiao et~al.(2023)Xiao, Lin, Seznec, Wu, Demouth, and Han}]{xiao2022smoothquant}
Guangxuan Xiao, Ji~Lin, Mickael Seznec, Hao Wu, Julien Demouth, and Song Han. 2023.
\newblock Smoothquant: Accurate and efficient post-training quantization for large language models.
\newblock In \emph{International Conference on Machine Learning}.

\bibitem[{Yang et~al.(2025)Yang, Yang, Zhang, Hui, Zheng, Yu, Li, Liu, Huang, Wei et~al.}]{qwen2025qwen25technicalreport}
An~Yang, Baosong Yang, Beichen Zhang, Binyuan Hui, Bo~Zheng, Bowen Yu, Chengyuan Li, Dayiheng Liu, Fei Huang, Haoran Wei, et~al. 2025.
\newblock \href {http://arxiv.org/abs/2412.15115} {Qwen2.5 technical report}.

\bibitem[{You et~al.(2023)You, Sun, Shi, Yu, Zhao, Zhang, Li, Li, and Lin}]{you2023vitcod}
Haoran You, Zhanyi Sun, Huihong Shi, Zhongzhi Yu, Yang Zhao, Yongan Zhang, Chaojian Li, Baopu Li, and Yingyan Lin. 2023.
\newblock Vitcod: Vision transformer acceleration via dedicated algorithm and accelerator co-design.
\newblock In \emph{2023 IEEE International Symposium on High-Performance Computer Architecture (HPCA)}, pages 273--286. IEEE.

\bibitem[{Yuan et~al.(2024)Yuan, Shang, and Dong}]{yuanpb}
Zhihang Yuan, Yuzhang Shang, and Zhen Dong. 2024.
\newblock Pb-llm: Partially binarized large language models.
\newblock In \emph{The Twelfth International Conference on Learning Representations}.

\bibitem[{Yuan et~al.(2023)Yuan, Shang, Song, Wu, Yan, and Sun}]{yuan2023asvd}
Zhihang Yuan, Yuzhang Shang, Yue Song, Qiang Wu, Yan Yan, and Guangyu Sun. 2023.
\newblock Asvd: Activation-aware singular value decomposition for compressing large language models.
\newblock \emph{arXiv preprint arXiv:2312.05821}.

\bibitem[{Zellers et~al.(2019)Zellers, Holtzman, Bisk, Farhadi, and Choi}]{zellers2019hellaswag}
Rowan Zellers, Ari Holtzman, Yonatan Bisk, Ali Farhadi, and Yejin Choi. 2019.
\newblock Hellaswag: Can a machine really finish your sentence?
\newblock In \emph{Proceedings of the 57th Annual Meeting of the Association for Computational Linguistics}.

\bibitem[{Zhang et~al.(2024{\natexlab{a}})Zhang, Chen, Shen, Yang, Ou, Yu, and Zhuang}]{zhang2023loraprune}
Mingyang Zhang, Hao Chen, Chunhua Shen, Zhen Yang, Linlin Ou, Xinyi Yu, and Bohan Zhuang. 2024{\natexlab{a}}.
\newblock \href {http://arxiv.org/abs/2305.18403} {Loraprune: Structured pruning meets low-rank parameter-efficient fine-tuning}.

\bibitem[{Zhang et~al.(2022)Zhang, Roller, Goyal, Artetxe, Chen, Chen, Dewan, Diab, Li, Lin et~al.}]{zhang2022opt}
Susan Zhang, Stephen Roller, Naman Goyal, Mikel Artetxe, Moya Chen, Shuohui Chen, Christopher Dewan, Mona Diab, Xian Li, Xi~Victoria Lin, et~al. 2022.
\newblock \href {http://arxiv.org/abs/2205.01068} {{OPT}: Open pre-trained transformer language models}.

\bibitem[{Zhang et~al.(2024{\natexlab{b}})Zhang, Bai, Lin, Zhao, Hou, and Cannistraci}]{zhang2024plug}
Yingtao Zhang, Haoli Bai, Haokun Lin, Jialin Zhao, Lu~Hou, and Carlo~Vittorio Cannistraci. 2024{\natexlab{b}}.
\newblock Plug-and-play: An efficient post-training pruning method for large language models.
\newblock In \emph{The Twelfth International Conference on Learning Representations}.

\bibitem[{Zhang et~al.(2024{\natexlab{c}})Zhang, Zhao, Lin, Sun, Yao, Han, Tanner, Liu, and Ji}]{zhang2024dynamic}
Yuxin Zhang, Lirui Zhao, Mingbao Lin, Yunyun Sun, Yiwu Yao, Xingjia Han, Jared Tanner, Shiwei Liu, and Rongrong Ji. 2024{\natexlab{c}}.
\newblock \href {http://arxiv.org/abs/2310.08915} {Dynamic sparse no training: Training-free fine-tuning for sparse llms}.

\bibitem[{Zhou and Tao(2011)}]{zhou2011godec}
Tianyi Zhou and Dacheng Tao. 2011.
\newblock Godec: Randomized low-rank \& sparse matrix decomposition in noisy case.
\newblock In \emph{Proceedings of the 28th International Conference on Machine Learning, ICML 2011}.

\end{thebibliography}

\appendix

\section{Convergence Analysis}
\label{app:convergence}

Building upon Optimal Brain Surgeon (OBS)~\cite{hassibi1993obs}, with extensions in SparseGPT~\cite{frantar2023sparsegpt} and GPTQ~\cite{frantar2023gptq}, the element-wise perturbation at $(i,j)$ induces quadratic error:
\begin{equation}
\delta_{i,j} = \frac{\Delta W_{ij}^2}{[H^{-1}]_{jj}^2}
\approx \|\Delta W \| \cdot \| X_j \|_2
\end{equation}

To jointly optimize the \textbf{low-rank} ($L$) and \textbf{sparse} ($S$) matrices:
\begin{equation}
\arg\min \| (W - L - S) X \|_F
\approx \| W - L - S \| \cdot \| X_j \|_2
\end{equation}

We solve $L$ and $S$ iteratively (Eq. \ref{eq:prune1} and Eq. \ref{eq:svd} in main text), defining optimization losses:
\begin{align*}
E^1_t &\approx \| (W - L_t - S_{t-1}) \| \cdot \| X_j \|_2 \\
E^2_t &\approx \| (W - L_t - S_t) \| \cdot \| X_j \|_2
\end{align*}

Global optimality of $S_t$ and $L_{t+1}$ ensures:
\begin{align}
E^1_t &\geq E^2_t \label{eq:dec1} \\
E^2_t &\geq E^1_{t+1} \label{eq:dec2}
\end{align}

Thus the quadratic error $\| (W - L - S) \| \cdot \| X_j \|_2$ decreases monotonically:
\begin{equation}
E^1_1 \geq E^2_1 \geq E^1_2 \geq \cdots \geq E^1_t \geq E^2_t \geq E^1_{t+1} \geq \cdots
\end{equation}

Complementing this theoretical framework, Figure \ref{fig:iterative_loss} (main text) shows monotonic error reduction across layers, with >90\% convergence within 40 iterations.

\section{Detailed Experimental Settings} \label{sec:appendix:set}

\subsection{Setup.} It is worth noting that our synergistic optimization method, is a simple and efficient way to run on consumer-grade graphics cards, where the largest computing resource is needed in fine-tuning schemes. The calibration dataset used in the experiments is the same as Wanda, sampled from the first slice of the C4~\cite{raffel2020c4} training dataset, containing 128 sequences with 2048 tokens each, which reflects the reality of the baseline approach. We use high quality instruction dataset Stanford Alpaca~\cite{taori2023alpaca} dataset for fine-tuning the compressed models.

\subsection{Models.}Our evaluation primarily focuses on leading open-source LLM families, including the LLaMA series and Qwen2.5 models. Specifically, we validate our method across multiple architectures and scales: LLaMA-7B/13B, LLaMA2-7B/13B, LLaMA3-8B/70B, and Qwen2.5-7B/14B. The empirical results demonstrate that our approach achieves consistent performance improvements regardless of model size or architecture.

\subsection{Evaluation.}Experiments evaluated on the WikiText-2~\cite{merity2016wiki2}, C4 datasets for perplexity ($PPL$) validation. To explore the model's capabilities in depth, we follow previous methods to perform zero-shot task classification with the help of the lm-eval~\cite{eval-harness} library on datasets including BoolQ~\cite{clark-etal-2019-boolq}, PIQA~\cite{Bisk2020piqa}, HellaSwag~\cite{zellers2019hellaswag}, WinoGrande~\cite{ai2:winogrande}, ARC-easy~\cite{allenai:arc}, and ARC-challenge~\cite{allenai:arc}. The licenses for the datasets and models used in this paper are as follows:

\begin{itemize}
  \item \textbf{WikiText-2}: Creative Commons Attribution-ShareAlike.
  \item \textbf{C4}: Apache License 2.0.
  \item \textbf{BoolQ}: Creative Commons Attribution-ShareAlike 3.0 (CC BY-SA 3.0).
  \item \textbf{PIQA}: MIT License.
  \item \textbf{HellaSwag}: MIT License.
  \item \textbf{WinoGrande}: Creative Commons Attribution 4.0 (CC BY 4.0).
  \item \textbf{ARC-easy / ARC-challenge}: Creative Commons Attribution-ShareAlike 4.0 (CC BY-SA 4.0).
  \item \textbf{LLaMA1}: Non-commercial research license; 
  \item \textbf{LLaMA2}: Meta Llama 2 Community License;
  \item \textbf{LLaMA3}: Meta Llama 3 Community License; 
  \item \textbf{Qwen2.5}: Apache License 2.0; 
\end{itemize}

All datasets and models were utilized in accordance with their respective licenses.

\subsection{Baselines.} We have meticulously reproduced several established methodologies to serve as benchmarks: (1) SparseGPT, which ingeniously reframes the task of model pruning in LLMs as a sequential sparse regression challenge, subsequently updating the unpruned weights. (2) Wanda, a method that approximates the SparseGPT pruning metric using the product of the magnitude of weights and L2 normalization based on input activation, performing only weight pruning. (3) DSNoT, a dynamic pruning technique that expands upon the sparse methodologies like Wanda, engaging in iterative processes of weight pruning and growth, which can be seen as an iterative optimization algorithm of sparse plus sparse. (4) SVD-LLM, a novel SVD-based LLM compression method, addresses the limitations of existing SVD approaches by incorporating a truncation-aware data whitening strategy that directly maps singular values to compression loss, thereby demonstrating superior performance compared to previous SVD compression methods~\cite{yuan2023asvd,hsu2022language}.

\section{Detailed Simulated ViTCoD Accelerator}\label{sec::vitcod}

ViTCoD~\cite{you2023vitcod} is an innovative framework for algorithm and hardware co-design. It effectively reduces the demand for on-chip cache and the frequency of input matrix loading by spatially tiling sparse and dense matrices along specific dimensions and accumulating intermediate results. During the computation, VITCoD divides the input matrices into smaller blocks and transfers them to memory buffers, then intelligently assigns computation tasks to either the Denser Engine or the Sparser Engine based on the sparsity of the matrix columns. The partial results computed by the Denser Engine are then transferred to the Sparser Engine for accumulation. This strategy not only enhances the reuse rate of input matrices and reduces the need for on-chip buffers but also optimizes the utilization of processing elements by reasonably distributing computation tasks, thereby improving overall computational performance.

\section{Detailed Zero-shot Task Performance}
\label{sec:detail_performance}

We evaluated a series of zero-shot learning tasks, as shown in Tables ~\ref{tab:language_modelling}. We present detailed task performance metrics in Tables ~\ref{tab:language_modelling_detailed},
providing a comprehensive understanding of the zero-shot capabilities of the related models.

\section{Detailed Ablation Study}\label{sec:method:ablation}

\subsection{Different Ranks.}With a fixed compression ratio of 50\%, an in-depth analysis of the effects of sparse and low-rank parameter assignments on LLaMA2-7B model are provided. As demonstrated in Table~\ref{tab:ablate_number_of_rank}, the model performance improves when the rank is increased from 32 to 128; however, after 128, the performance starts to decrease. Therefore, 128 is chosen as the optimal compromise point for parameter allocation to balance model performance, which is significantly better than pure pruning methods (rank=0) or pure low-rank methods (rank=1296). The results of this study not only highlight the need to balance pruning and low rank in model design, but also provide valuable reference for the development of algorithms to find the optimal combination.

\begin{table}[ht!]
 \centering
    \renewcommand{\arraystretch}{1}
    \small
    \begin{tabular}{cccccccc}\toprule
    Dataset   & \begin{tabular}[c]{@{}c@{}}r=0\\\end{tabular} & r=64 & \textbf{r=128} & r=256 &r=1296 \\ 
    \midrule
    Wiki2 & 6.92     & 6.72 & \cellcolor[HTML]{EFEFEF}\textbf{6.61}  & 6.70    &1.02e4  \\ 
    C4        & 9.24    & 8.97 & \cellcolor[HTML]{EFEFEF}\textbf{8.87}   & 9.03 &1.85e4 \\
    \bottomrule
    \end{tabular}
\caption{Perplexity results for LLaMA2-7B at 50\% compression with different number of rank. When r=1296, this is a pure low-rank approximation with 0\% sparsity; in contrast, when r=0, this corresponds to a pure pruning approach with 50\% sparsity.}
\label{tab:ablate_number_of_rank}
\end{table}

\subsection{Preserving Most Important Weights.}We explore the effects of preserving the most important weights prior to synergistic optimization. The findings are detailed in the Table~\ref{tab:retain weight percent}. The results show that incorporating this retention ratio at a 1\% level leads to the best improvement in performance, while at a 10\% level, the performance declines sharply. Additionally, it is important to highlight that these 1\% weights can be seamlessly integrated into the sparse part, incurring no extra structural cost.

\begin{table}[h!]
    \centering 
     \renewcommand{\arraystretch}{1}
    \small
    \begin{tabular}{cccc}
    \toprule
    \multicolumn{1}{l}{Models}  & \multicolumn{1}{c}{\begin{tabular}[c]{@{}c@{}}Preserved\\ Ratio\end{tabular}} & \multicolumn{1}{c}{Wiki2} & \multicolumn{1}{c}{C4} \\ 
    \midrule
    \multirow{4}{*}{LLaMA2-7B}  & 0\%                              & 6.71                          & 8.97                   \\
                                &\cellcolor[HTML]{EFEFEF} \textbf{ 1\%}                              &\cellcolor[HTML]{EFEFEF}\textbf{6.61}                         & \cellcolor[HTML]{EFEFEF}\textbf{8.87}                   \\
                                & 3\%                              & 6.63                          & \textbf{8.87}                   \\
                                & 10\%                             & 6.70                           & 8.99                   \\ 
    \midrule
    \multirow{4}{*}{LLaMA2-13B} & 0\%                              & 8.10                              &5.84                     \\                                        
                                & \cellcolor[HTML]{EFEFEF} \textbf{1\% }                             &\cellcolor[HTML]{EFEFEF} \textbf{8.02}                        &\cellcolor[HTML]{EFEFEF}\textbf{5.79}                     \\
                                & 3\%                              & 8.03                            & 5.80                       \\
                                & 10\%                             & 8.06                              &5.82                        \\ 
    \bottomrule
    \end{tabular}
    \caption{Perplexity results for LLaMA2-7B and LLaMA2-13B at 50\% compression with retaining different proportions of the most importance weights.}
    \label{tab:retain weight percent}
    \vspace{-1mm}
\end{table}

\begin{table*}[h]
    \centering
    \renewcommand{\arraystretch}{1}
\resizebox{\linewidth}{!}{
\begin{tabular}{ccccccccccc}
\hline
\multicolumn{2}{c}{Method}       & PIQA  & Boolq & HellaS & Wino  & ARC-e & ARC-c & Ave            & Wiki2         & C4            \\ \hline
\multirow{6}{*}{Wanda} & Overall & 76.24 & 76.14 & 52.72  & 67.97 & 72.14 & 39.00 & 64.04{\small ±0.10}         & 6.92{\small ±0.01}          & 9.23{\small ±0.01}          \\ \cline{2-11} 
                       & Seed\_0 & 76.71 & 76.60 & 52.56  & 68.43 & 72.18 & 38.31 & 64.13          & 6.92          & 9.24          \\
                       & Seed\_1 & 76.16 & 75.66 & 52.62  & 68.03 & 72.47 & 39.51 & 64.08          & 6.91          & 9.25          \\
                       & Seed\_2 & 76.06 & 76.42 & 52.75  & 67.88 & 71.72 & 39.51 & 64.06          & 6.91          & 9.23          \\
                       & Seed\_3 & 76.11 & 76.02 & 52.70  & 68.19 & 72.26 & 38.99 & 64.05          & 6.93          & 9.23          \\
                       & Seed\_4 & 76.17 & 75.99 & 52.99  & 67.32 & 72.05 & 38.66 & 63.86          & 6.94          & 9.22          \\ \hline
\multirow{6}{*}{DSnoT} & Overall & 75.94 & 74.04 & 54.89  & 64.09 & 64.91 & 44.86 & 63.12{\small ±0.09}          & 6.85{\small ±0.02}          & 9.12{\small ±0.01}          \\ \cline{2-11} 
                       & Seed\_0 & 76.28 & 73.58 & 52.01  & 66.93 & 71.68 & 38.82 & 63.22          & 6.83          & 9.13          \\
                       & Seed\_1 & 75.95 & 74.77 & 51.84  & 67.32 & 71.21 & 37.71 & 63.13          & 6.85          & 9.11          \\
                       & Seed\_2 & 75.90 & 74.46 & 51.91  & 66.77 & 71.25 & 38.05 & 63.06          & 6.86          & 9.11          \\
                       & Seed\_3 & 75.73 & 73.58 & 51.84  & 67.01 & 71.67 & 38.22 & 63.01          & 6.87          & 9.12          \\
                       & Seed\_4 & 75.84 & 73.82 & 51.94  & 67.32 & 71.59 & 38.65 & 63.19          & 6.84          & 9.11          \\ \hline
\multirow{6}{*}{Ours}  & Overall & \cellcolor[HTML]{EFEFEF}\textbf{77.15} & \cellcolor[HTML]{EFEFEF}\textbf{76.93} & \cellcolor[HTML]{EFEFEF}\textbf{53.89} & \cellcolor[HTML]{EFEFEF}\textbf{68.40} & \cellcolor[HTML]{EFEFEF}\textbf{73.94} & \cellcolor[HTML]{EFEFEF}\textbf{41.19} & \cellcolor[HTML]{EFEFEF}\textbf{65.25{\small ±0.10}} & \cellcolor[HTML]{EFEFEF}\textbf{6.62{\small ±0.02}} & \cellcolor[HTML]{EFEFEF}\textbf{8.87{\small ±0.00}} \\ \cline{2-11} 
                       & Seed\_0 & 76.55 & 77.68 & 53.81  & 67.32 & 74.41 & 40.96 & 65.12          & 6.61          & 8.87          \\
                       & Seed\_1 & 77.47 & 76.33 & 53.89  & 68.82 & 73.93 & 41.88 & 65.39          & 6.61          & 8.87          \\
                       & Seed\_2 & 77.21 & 77.73 & 53.99  & 68.35 & 73.19 & 40.70 & 65.20           & 6.64          & 8.87          \\
                       & Seed\_3 & 77.42 & 77.83 & 53.87  & 69.46 & 73.15 & 40.10 & 65.31          & 6.59          & 8.87          \\
                       & Seed\_4 & 77.09 & 75.08 & 53.89  & 68.03 & 75.04 & 42.32 & 65.24          & 6.64          & 8.87          \\ \hline
\end{tabular}}
    \caption{Accuracy on zero-shot tasks and language modeling performance ($PPL\downarrow$) for LLaMA2-7B at 50\% compression rate across different pruning methods (mean±std over 5 random seeds).} 
    \label{tab:language_modelling_detailed_seed}
    \vspace{-1mm}
\end{table*}

\begin{table*}[t]
    \centering
    \renewcommand{\arraystretch}{1.2}
    \small
    \begin{tabular}{ccccccccc|cc}
\hline
Method &Conference                   & PIQA                                   & BoolQ                                  & HellaS                                 & Wino                                   & ARC-e                                  & ARC-c                                  & Ave  & Wiki2 & C4                                  \\ \hline
RIA                       &ICLR2024& 76.11                                  & 75.57                                  & 52.21                                  & 67.48                                  & 71.51                                  & \textbf{38.39}                         & 63.55 &6.81&9.11                                 \\
\textbf{RIA+ours}       &  & \cellcolor[HTML]{EFEFEF}\textbf{76.93} & \cellcolor[HTML]{EFEFEF}\textbf{76.12} & \cellcolor[HTML]{EFEFEF}\textbf{52.95} & \cellcolor[HTML]{EFEFEF}\textbf{69.61} & \cellcolor[HTML]{EFEFEF}\textbf{72.81} & \cellcolor[HTML]{EFEFEF}38.14          & \cellcolor[HTML]{EFEFEF}\textbf{64.42}& \cellcolor[HTML]{EFEFEF}\textbf{6.54}& \cellcolor[HTML]{EFEFEF}\textbf{8.77} \\ \hline
ALPS               &NIPS2024& 76.22                                  & 75.37                         & 53.12                                  & 68.21                                  & 72.61                                  & 41.21                                  & 64.46 &6.87&9.01                                 \\
\textbf{ALPS+ours} && \cellcolor[HTML]{EFEFEF}\textbf{76.44} & \cellcolor[HTML]{EFEFEF}\textbf{76.64}         & \cellcolor[HTML]{EFEFEF}\textbf{53.87} & \cellcolor[HTML]{EFEFEF}\textbf{69.22}  & \cellcolor[HTML]{EFEFEF}\textbf{73.19} & \cellcolor[HTML]{EFEFEF}\textbf{41.32} & \cellcolor[HTML]{EFEFEF}\textbf{65.11}& \cellcolor[HTML]{EFEFEF}\textbf{6.60}& \cellcolor[HTML]{EFEFEF}\textbf{8.73} \\\hline

Pruner-Zero  &ICML2024             & 75.90                                  & \textbf{74.13}                         & 51.16                                  & 67.01                                  & 71.17                                  & 37.28                                  & 62.78&6.61&9.23                                  \\
\textbf{Pruner-Zero+ours} & &\cellcolor[HTML]{EFEFEF}\textbf{76.17} & \cellcolor[HTML]{EFEFEF}73.88          & \cellcolor[HTML]{EFEFEF}\textbf{51.41} & \cellcolor[HTML]{EFEFEF}\textbf{69.16}  & \cellcolor[HTML]{EFEFEF}\textbf{72.73} & \cellcolor[HTML]{EFEFEF}\textbf{39.59} & \cellcolor[HTML]{EFEFEF}\textbf{63.82}& \cellcolor[HTML]{EFEFEF}\textbf{6.45}& \cellcolor[HTML]{EFEFEF}\textbf{8.88} \\
\hline
\end{tabular}
    \caption{Accuracy on zero-shot tasks and language modeling performance ($PPL$) for LLaMA2-7B of 50\% compression rate across different pruning methods.} 
    \label{tab:ablate_zero}
    \vspace{-1mm}
\end{table*}

\subsection{Random Seeds.} To address potential concerns regarding the reproducibility of performance differences, we conducted a comprehensive robustness analysis across five distinct random seeds (0-4) under identical hyperparameter configurations. Our method demonstrates exceptional stability and robustness, maintaining consistent superiority over baseline approaches despite varying initialization conditions. As evidenced in Table ~\ref{tab:language_modelling_detailed_seed}, SSLC achieves statistically significant improvements across all evaluation tasks, with performance variances remaining below 0.02 standard deviation for both our method and competitors on stable benchmarks like C4 and WikiText-2, while the average accuracy on zero-shot tasks exhibit $\sigma\approx0.1$ across all compared methods.

\subsection{SSLC with Other LLM Pruning Methods.}\label{sec:sslc_with_other_method}Our framework establishes new capabilities for model compression by simultaneously enhancing both task performance and intrinsic language modeling across diverse pruning methods. The results in Table~\ref{tab:ablate_zero} demonstrate that, as a universal plugin, it consistently improves accuracy on reasoning benchmarks (+0.7-1.0\% average) while reducing perplexity across all baselines. 

\section{Potential Risks}
While our method effectively maintains model performance at moderate sparsity (e.g., 50\%), excessive pruning introduces significant performance degradation risks. This underscores a critical limitation of post-training pruning: aggressive sparsification cannot be fully remedied by fine-tuning alone, potentially compromising model reliability in high-sparsity scenarios.

\begin{table*}[thb]
    \centering
    \renewcommand{\arraystretch}{1}
\resizebox{\linewidth}{!}{
\begin{tabular}{cccccccccc}
\hline
Model                        & Method    & Type  & PIQA  & BoolQ & HellaS & Wino  & ARC-e & ARC-c & Ave            \\ \hline
\multirow{5}{*}{LLaMA-7B}    & Dense     & -     & 78.67 & 75.08 & 56.94  & 70.01 & 75.25 & 41.89 & 66.31          \\ \cline{2-10} 
                             & SparseGPT & S     & 76.39 & 72.97 & 51.41  & 69.38 & 71.30 & 37.29 & 63.12          \\
                             & Wanda     & S     & 76.04 & 71.62 & 52.48  & 68.74 & 70.75 & 37.03 & 62.77          \\
                             & DSnoT     & S     & 76.01 & 73.09 & 52.87  & 67.40 & 70.95 & 37.12 & 62.91          \\
                             & Ours      & S+LRA & 76.33 & 74.95 & 52.97  & 68.82 & 71.68 & 36.77 & \textbf{63.59} \\ \hline
\multirow{5}{*}{LLaMA2-7B}   & Dense     & -     & 78.07 & 77.71 & 57.14  & 68.90 & 76.35 & 43.60 & 66.96          \\ \cline{2-10} 
                             & SparseGPT & S     & 76.17 & 76.02 & 52.81  & 68.67 & 71.63 & 36.95 & 63.71          \\
                             & Wanda     & S     & 76.71 & 76.60 & 52.56  & 68.43 & 72.18 & 38.31 & 64.13          \\
                             & DSnoT     & S     & 76.28 & 73.58 & 52.01  & 66.93 & 71.68 & 38.82 & 63.22          \\
                             & Ours      & S+LRA & 77.09 & 75.08 & 53.89  & 68.03 & 75.04 & 42.32 & \textbf{65.24} \\ \hline
\multirow{5}{*}{LLaMA3-8B}   & Dense     & -     & 80.14 & 82.08 & 60.02  & 73.64 & 81.40 & 51.19 & 71.41          \\ \cline{2-10} 
                             & SparseGPT & S     & 76.22 & 78.13 & 53.65  & 71.43 & 72.43 & 41.21 & 65.51          \\ 
                             & Wanda     & S     & 75.90 & 79.54 & 51.41  & 70.96 & 73.23 & 41.64 & 65.44          \\
                             & DSnoT     & S     & 75.52 & 79.05 & 51.51  & 69.38 & 73.15 & 40.87 & 64.91          \\
                             & Ours      & S+LRA & 76.39 & 78.57 & 53.18  & 70.64 & 74.71 & 42.32 & \textbf{65.97} \\ \hline
\multirow{5}{*}{LLaMA-13B}   & Dense     & -     & 79.16 & 77.89 & 59.93  & 72.69 & 77.36 & 46.42 & 68.91          \\ \cline{2-10} 
                             & SparseGPT & S     & 78.35 & 76.85 & 54.88  & 71.35 & 72.47 & 41.98 & 65.98          \\
                             & Wanda     & S     & 77.42 & 76.67 & 55.82  & 72.06 & 74.07 & 43.43 & 66.58          \\
                             & DSnoT     & S     & 77.48 & 76.45 & 55.68  & 71.19 & 73.78 & 43.86 & 66.41          \\
                             & Ours      & S+LRA & 78.29 & 75.59 & 56.48  & 70.96 & 75.21 & 45.39 & \textbf{66.99} \\ \hline
\multirow{5}{*}{LLaMA2-13B}  & Dense     & -     & 79.05 & 80.55 & 60.06  & 72.14 & 79.42 & 48.46 & 69.95          \\ \cline{2-10} 
                             & SparseGPT & S     & 77.69 & 81.41 & 55.93  & 71.59 & 74.66 & 42.06 & 67.22          \\
                             & Wanda     & S     & 78.41 & 81.19 & 57.09  & 71.35 & 76.98 & 43.00 & 68.01          \\
                             & DSnoT     & S     & 77.91 & 80.70 & 57.02  & 71.72 & 76.64 & 42.58 & 67.78          \\
                             & Ours      & S+LRA & 78.24 & 81.22 & 57.40  & 71.43 & 76.94 & 46.08 & \textbf{68.55} \\ \hline
\multirow{5}{*}{LLaMA3-70B}  & Dense     & -     & 82.32 & 85.26 & 66.38  & 80.51 & 86.86 & 60.15 & 76.91          \\ \cline{2-10} 
                             & SparseGPT & S     & 81.77 & 84.95 & 62.81  & 76.80 & 83.25 & 55.55 & 74.19          \\
                             & Wanda     & S     & 81.07 & 85.32 & 62.52  & 79.42 & 82.95 & 55.03 & 74.39          \\
                             & DSnoT     & S     & 81.56 & 84.74 & 63.13  & 77.58 & 83.25 & 55.38 & 74.27          \\
                             & Ours      & S+LRA & 82.26 & 85.17 & 63.16  & 78.37 & 83.79 & 55.97 & \textbf{74.79} \\ \hline
\multirow{5}{*}{Qwen2.5-7B}  & Dense     & -     & 78.51 & 84.52 & 72.77  & 60.01 & 80.56 & 48.63 & 70.83          \\ \cline{2-10} 
                             & SparseGPT     & S     & 77.42 & 83.09 & 71.11  & 54.63 & 76.60 & 44.03 & 67.81          \\
                             &Wanda  & S     & 77.15 & 83.03 & 70.24  & 53.07 & 75.59 & 41.12 & 66.70          \\
                             & DSnoT     & S     & 77.04 & 83.21 & 70.95  & 52.96 & 75.72 & 41.46 & 66.89          \\
                             & Ours      & S+LRA & 77.81 & 83.30 & 71.35  & 54.44 & 79.00 & 46.16 & \textbf{68.68} \\ \hline
\multirow{5}{*}{Qwen2.5-14B} & Dense     & -     & 81.12 & 85.54 & 75.37  & 63.39 & 82.37 & 55.80 & 73.93          \\ \cline{2-10} 
                             & SparseGPT & S     & 79.00 & 85.69 & 73.24  & 57.25 & 80.85 & 51.11 & 71.19          \\
                             & Wanda     & S     & 78.78 & 85.69 & 73.32  & 57.25 & 80.93 & 50.94 & 71.15          \\
                             & DSnoT     & S     & 78.82 & 85.60 & 73.32  & 57.70 & 80.89 & 51.02 & 71.23          \\
                             & Ours      & S+LRA & 79.76 & 84.74 & 73.72  & 58.12 & 81.94 & 53.32 & \textbf{71.93} \\ \hline
\end{tabular}}
    \caption{Accuracy for zero-shot tasks on LLaMA and Qwen2.5 models of 50\% compression rate with different pruning methods.} 
    \label{tab:language_modelling_detailed}
    \vspace{-1mm}
\end{table*}

\end{document}